%% file: ms.tex
\documentclass{article}

\usepackage{microtype}
\usepackage{graphicx}
\usepackage{booktabs} 

\usepackage{hyperref}

\usepackage[accepted]{icml2018}

\graphicspath{ {img/} }    
\usepackage{subcaption} 

\usepackage{amssymb}
\usepackage{amsmath}
\usepackage{amsfonts}

\usepackage{algorithm}
\usepackage{algorithmic}

\newcommand{\R}{{\mathbb R}}

\newcommand{\T}{{\mathbb T}}
\newcommand{\X}{{\mathcal X}}
\newcommand{\Y}{{\mathcal Y}}

\icmltitlerunning{Deep Generative Dual Memory Network for Continual Learning}

\begin{document}

\twocolumn[
\icmltitle{Deep Generative Dual Memory Network for Continual Learning}




\begin{icmlauthorlist}
\icmlauthor{Nitin Kamra}{usc}
\icmlauthor{Umang Gupta}{usc}
\icmlauthor{Yan Liu}{usc}
\end{icmlauthorlist}

\icmlaffiliation{usc}{Department of Computer Science, University of Southern California, Los Angeles, CA, USA}

\icmlcorrespondingauthor{Nitin Kamra}{nkamra@usc.edu}

\icmlkeywords{Continual Learning, Incremental Learning, Lifelong Learning, Catastrophic Forgetting, Multitask Learning}

\vskip 0.3in
]



\printAffiliationsAndNotice{}  

\input{src/abstract.tex}

\input{src/intro.tex}

\input{src/smtl.tex}

\input{src/dgdmn.tex}

\input{src/exp.tex}

\input{src/anal.tex}

\input{src/conc.tex}


\newpage

\bibliography{bibfile}
\bibliographystyle{icml2018}

\clearpage

\input{src/appendixA.tex}
\input{src/appendixB.tex}

\end{document}

%% file: src/abstract.tex
\begin{abstract}
\label{abstract}

Despite advances in deep learning, neural networks can only learn multiple tasks when trained on them jointly. When tasks arrive sequentially, they lose performance on previously learnt tasks. This phenomenon called catastrophic forgetting is a fundamental challenge to overcome before neural networks can learn continually from incoming data. In this work, we derive inspiration from human memory to develop an architecture capable of learning continuously from sequentially incoming tasks, while averting catastrophic forgetting. Specifically, our contributions are: (i) a dual memory architecture emulating the complementary learning systems (hippocampus and the neocortex) in the human brain, (ii) memory consolidation via generative replay of past experiences, (iii) demonstrating advantages of generative replay and dual memories via experiments, and (iv) improved performance retention on challenging tasks even for low capacity models. Our architecture displays many characteristics of the mammalian memory and provides insights on the connection between sleep and learning.

\end{abstract}

%% file: src/intro.tex
\section{Introduction}
\label{sec:intro}

Many machine learning models, when trained sequentially on tasks, forget how to perform previously learnt tasks. This phenomenon, called \emph{catastrophic forgetting} is an important challenge to overcome in order to enable systems to learn continuously.
In the early stages of investigation, \citet{mccloskey1989catastrophic} suggested the underlying cause of forgetting to be the distributed shared representation of tasks via network weights.
Subsequent works attempted to reduce representational overlap between input representations via activation sharpening algorithms~\citep{kortge1990episodic}, orthogonal recoding of inputs~\citep{lewandowsky1991gradual} and orthogonal activations at all hidden layers~\citep{mcrae1993catastrophic, french1994dynamically}.
Recently, activations like maxout and dropout~\citep{goodfellow2013empirical} and local winner-takes-all~\citep{srivastava2013compete} have been explored to create sparsified feature representations.
But, natural cognitive systems e.g. mammalian brains are also connectionist in nature and yet they only undergo gradual systematic forgetting. Frequently and recently encountered tasks tend to survive much longer in memory, while those rarely encountered are slowly forgotten. Hence shared representations may not be the root cause of the problem.
More recent approaches have targeted slowing down learning on network weights which are important for previously learnt tasks. \citet{kirkpatrick2017overcoming} have used a fisher information matrix based regularizer to slow down learning on network weights which correlate with previously acquired knowledge. \citet{zenke2017continual} have employed path integrals of loss-derivatives to slow down learning on weights important for the previous tasks. 
Progressive neural networks~\citep{rusu2016progressive} and Pathnets~\citep{fernando2017pathnet} directly freeze important pathways in neural networks, which eliminates forgetting altogether but requires growing the network after each task and can cause the architecture complexity to grow with the number of tasks. \citet{li2017learning} have evaluated freezing weights in earlier layers of a network and fine tuning the rest for multiple tasks. These methods outperform sparse representations but may not be explicitly targeting the cause of catastrophic forgetting.

An important assumption for successful gradient-based learning is to observe \emph{iid} samples from the joint distribution of all tasks to be learnt. Since sequential learning systems violate this assumption, catastrophic forgetting is inevitable. So a direct approach would be to store previously seen samples and replay them along with new samples in appropriate proportions to restore the \emph{iid} sampling assumption~\citep{lopez2017gradient}. This experience replay approach has been adopted by maintaining a fixed-size episodic memory of exemplars which are either directly replayed while learning e.g. in iCaRL~\citep{rebuffi2017icarl} or indirectly used to modify future gradient updates to the system e.g. in GEM~\citep{lopez2017gradient} to mitigate forgetting on previously seen tasks.
However, choosing to store samples from previous tasks is challenging since it requires determining how many samples need to be stored, which samples are most representative of a task, and which samples to discard as new tasks arrive~\citep{lucic2017coreset}.
We propose that this problem can be solved by maintaining a generative model over samples which would automatically provide the most frequently encountered samples from the distribution learnt so far. This is also feasible with limited total memory and avoids explicitly determining which and how many samples should be stored and/or discarded per task.
Previous non-generative approaches to experience replay e.g. pseudo-pattern rehearsal~\citep{robins2004sequential} have proposed to preserve neural networks' learnt mappings by uniformly sampling random inputs and their corresponding outputs from networks and replaying them along with new task samples. These approaches have only been tested in small binary input spaces and our experiments show that sampling random inputs in high-dimensional spaces (e.g. images) does not preserve the learnt mappings.

Neuroscientific evidence suggests that experience replay of patterns has also been observed in the human brain during sleep and waking rest~\citep{mcclelland1995there,o2010play}. Further, humans have evolved mechanisms to separately learn new incoming tasks and consolidate them with previous knowledge to avert catastrophic forgetting~\citep{mcclelland1995there,french1999catastrophic}.
The widely acknowledged complementary learning systems theory~\citep{mcclelland1995there,kumaran2016learning} suggests that this separation has been achieved in the human brain via evolution of two separate areas: (a) the neocortex, which is a long term memory specializing in consolidating new information with previous knowledge to gradually learn the joint structure of all tasks, and (b) the hippocampus, which acts as a temporary memory to rapidly learn new tasks and then slowly transfers the knowledge to neocortex after acquisition.

In this paper, we propose a dual-memory architecture for learning tasks sequentially while averting catastrophic forgetting.
Our model comprises of two generative models: a short-term memory (STM) to emulate the human hippocampal system and a long term memory (LTM) to emulate the neocortical learning system. The STM learns new tasks without interfering with previously learnt tasks in the LTM. The LTM stores all previously learnt tasks and aids the STM in learning tasks similar to previously seen tasks. During sleep/down-time, the STM generates and transfers samples of learnt tasks to the LTM. These are gradually consolidated with the LTM's knowledge base of previous tasks via generative replay.
Our model exploits the strengths of deep generative models, experience replay and complementary learning systems literature. We demonstrate its performance experimentally in averting catastrophic forgetting by sequentially learning multiple tasks. Moreover, our experiments shed light on some characteristics of human memory as observed in the psychology and neuroscience literature.

%% file: src/smtl.tex
\section{Problem Description}
\label{sec:smtl}

Formally, our problem setting is characterized by a set of tasks $\T$, to be learnt by a parameterized model. Note that we use the the phrase \emph{model} and \emph{neural network architecture} interchangeably. In this work, we mainly consider supervised learning tasks i.e.\ task $t \in \T$ has training samples: $\{X_t, Y_t\} = \{x_i^t, y_i^t\}_{i=1:N_t}$ for $x_i^t \in \X$ and $y_i^t \in \Y$, but our model easily generalizes to unsupervised learning settings.
Samples for each task are drawn \emph{iid} from an (unknown) data generating distribution $P_t$ associated with the task i.e. $\{x_i^t, y_i^t\} \sim P_t  \hspace{4pt} \forall i \in [N_t]$, but the distributions $\{P_t\}_{t \in \T}$ can be completely different from each other.
The tasks arrive sequentially and the total number of tasks $T=|\T|$ is not known a priori. Note that the full sequence of samples seen by the architecture is not sampled \emph{iid} from the joint distribution of all samples.
The architecture observes the task descriptor and the data $\{t, X_t, Y_t\}$ for each task while training sequentially. It can be evaluated at any time on a test sample $\{t, x_t\}$ to predict its label $y_t$ where $\{x_t, y_t\} \sim P_t$ after task $t$ has been observed.
Our goal is to learn these tasks sequentially while avoiding catastrophic forgetting and achieve a test accuracy close to that of a model which was jointly trained on all tasks.

\textbf{Finite memory}: We allow a limited storage for algorithms to store or generate samples while learning.The storage size is limited to $N_{max}$ and usually smaller than the total number of samples $\sum_{t=1}^{T} N_t$. Hence, just storing all training samples and reusing them is infeasible.

\textbf{Evaluation metrics}: After training on each task, we evaluate models on separate test sets for each task. This gives us a matrix $A \in \R^{T \times T}$ with $A_{i,j}$ being the test accuracy on task $j$ after training on task $i$. Following \cite{lopez2017gradient}, we evaluate algorithms on the following metrics --- Average accuracy (ACC) achieved across all tasks and Backward Transfer (BWT):
$$
    ACC = \frac{1}{T} \sum_{i=1}^{T} A_{T,i} \hspace{4pt} \bigg{|} \hspace{4pt}
    BWT = \frac{1}{T-1} \sum_{i=1}^{T-1} A_{T,i} - A_{i,i}
$$
Backward transfer (BWT) measures the influence of task $t$ on a previously learnt task $\tau$. This is generally negative since learning new tasks sequentially causes the model to lose performance on previous tasks. A large negative backward BWT represents catastrophic forgetting.
An ideal continual learning algorithm should achieve maximum ACC while having least negative (or positive) BWT.

%% file: src/dgdmn.tex
\begin{figure*}[th]
\vskip 0.2in
\begin{center}
\centerline{\includegraphics[width=\textwidth]{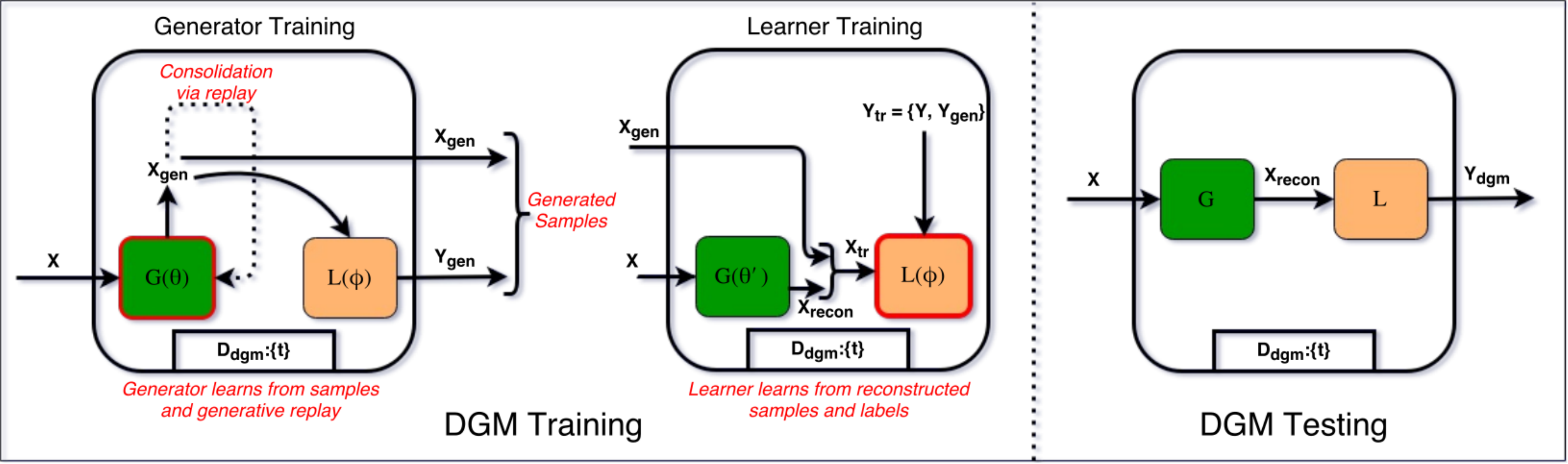}}
\caption{Deep Generative Replay to train a Deep Generative Memory}
\label{fig:dgm}
\end{center}
\vskip -0.2in
\end{figure*}

\section{Deep Generative Dual Memory Network}
\label{sec:dgdmn}

\subsection{Deep Generative Replay}
\label{subsec:dgr}

We present a generative experience replay algorithm to learn from sequentially arriving samples.
We first introduce a sub-model called the Deep Generative Memory (DGM)\footnote{We call this a memory because of its weights and learning capacity, not due to any recurrent connections.} with three elements: (i) a generative model (the generator $G$), (ii) a feedforward network (the learner $L$), and (iii) a dictionary ($D_{dgm}$) with task descriptors of learnt tasks and the number of times they were encountered.
Though most previous works~\cite{kirkpatrick2017overcoming,lopez2017gradient,zenke2017continual} and our algorithm involve usage of task descriptors $t$ in some form, our architecture also works when they are either unavailable, non-integral or just an inseparable part of the input $x_t$ (see Appendix A).
We choose variational autoencoder (VAE)~\citep{kingma2014vae} for the generator, since our generative model requires reconstruction capabilities (see section \ref{subsec:dmn}) but can also work with other kinds of generative models (see section \ref{sec:anal}).

\begin{algorithm}[b!]
    \caption{Deep Generative Replay}
    \label{alg:dgr}
\begin{algorithmic}[1]
    \STATE {\bfseries Input:} Current params and $age$ of DGM, new samples: $(X, Y)$, dictionary for new samples: $D_{tasks}$, minimum fraction: $\kappa$, memory capacity: $N_{max}$
    \STATE {\bfseries Output:} New parameters of DGM
    
    \COMMENT {Compute number of samples}
    \STATE $N_{tasks} = |X|$
    \STATE $N_{gen} = age$
    \IF {$|X| + age > N_{max}$}
        \STATE $\eta_{tasks} = \max \left( \kappa, \frac{|X|}{|X| + age} \right)$
        \STATE $N_{tasks} = \eta_{tasks} \times N_{max}$
        \STATE $N_{gen} = N_{max} - N_{tasks}$
    \ENDIF
    \STATE $N_{total} = N_{tasks} + N_{gen}$

    \COMMENT {Subsample $X, Y$ if needed}
    \IF {$N_{tasks} < |X|$}
        \STATE $X_{tasks}, Y_{tasks} =$ Draw $N_{tasks}$ samples from $X, Y$
    \ELSE
        \STATE $N_{tasks}, N_{gen} = |X|, N_{total} - |X|$
        \STATE $X_{tasks}, Y_{tasks} = X, Y$
    \ENDIF
    
    \COMMENT {Generate samples from previous tasks}
    \STATE $X_{gen} =$ Draw $N_{gen}$ samples from $G$
    \STATE $Y_{gen} = L(X_{gen})$
    \STATE $X_{tr}, Y_{tr} = \text{concat}(X_{tasks}, X_{gen}), \text{concat}(Y_{tasks}, Y_{gen})$
    
    \STATE Add task descriptors from $D_{tasks}$ to $D_{dgm}$
    \STATE $age = age + N_{total}$
    
    \COMMENT {Train DGM}
    \STATE Train generator $G$ on $X_{tr}$
    \STATE $X_{recon} =$ Reconstruct $X_{tasks}$ from generator $G$
    \STATE $X_{tr} = \text{concat}(X_{recon}, X_{gen})$
    \STATE Train learner $L$ on $(X_{tr}, Y_{tr})$
\end{algorithmic}
\end{algorithm}

We update a DGM with samples from (potentially multiple) new tasks using our algorithm Deep Generative Replay (DGR). The pseudocode is shown in algorithm \ref{alg:dgr} and visualized in figure \ref{fig:dgm}.
DGR essentially combines the new incoming samples $(X, Y)$ with its own generated samples from previous tasks and relearns jointly on these samples.
Given new incoming samples $(X, Y)$, DGR computes the fraction of samples to use from incoming samples $(\eta_{tasks})$ and the fraction to preserve from previous tasks $(\eta_{gen})$ according to the number of samples seen so far (i.e. $age$ of DGM).
If needed, the incoming samples are downsampled while still allocating at least a minimum fraction $\kappa$ of the memory to them (lines 3--16). This ensures that as the DGM saturates with tasks over time, new tasks are still learnt at the cost of gradually losing performance on the least recent previous tasks. This is synonymous to how learning slows down in humans as they age but they still continue to learn while forgetting old things gradually~\citep{french1999catastrophic}.
Next, DGR generates samples of previously learnt tasks $(X_{gen}, Y_{gen})$ using the generator and learner, transfers the task descriptors of samples in $(X, Y)$ to its own dictionary $D_{dgm}$ and updates its age (lines 17--21).
It then trains the generator on the total training samples $X_{tr}$, reconstructs the new samples via the trained generator as $X_{recon}$ (hence we use a VAE) and then trains the learner on resulting samples $X_{tr} = \text{concat}(X_{recon}, X_{gen})$ and their labels $Y_{tr}$ (lines 22--25). Doing this final reconstruction provides robustness to noise and occlusion (section \ref{subsec:resilience}).

Ideas similar to DGR have recently been proposed by \citet{mocanu2016rbm} and \citet{shin2017continual} independently, but they do not describe balancing new and generated samples and cannot recognize repeated tasks (section \ref{subsec:revision} in appendix A). Also generative replay without a dual memory architecture is costly to train (section \ref{subsec:longseq}) and a lack of reconstruction for new samples makes their representations less robust to noise and occlusions (section \ref{subsec:resilience}).

\begin{figure*}[th]
\vskip 0.2in
\begin{center}
\centerline{\includegraphics[width=\textwidth]{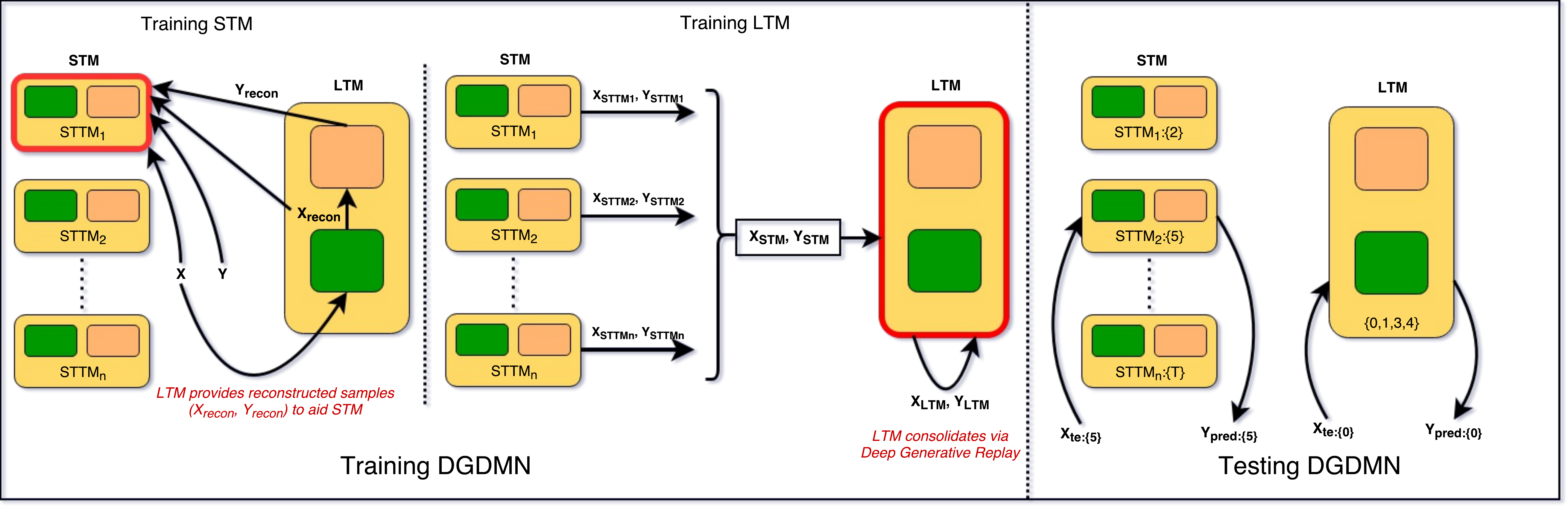}}
\caption{Deep Generative Dual Memory Network (DGDMN)}
\label{fig:dgdmn}
\end{center}
\vskip -0.2in
\end{figure*}

\subsection{Dual memory networks}
\label{subsec:dmn}

Though DGR is a continual learning algorithm on its own, our preliminary experiments showed that it is slow and inaccurate. To balance the conflicting requirements of quick acquisition of new tasks and performance retention on previously learnt tasks, we propose a dual memory network to combat forgetting.
Our architecture (DGDMN) shown in figure \ref{fig:dgdmn} comprises of a large DGM called the long-term memory (LTM) which stores information of all previously learnt tasks like the neocortex and a short-term memory (STM) which behaves similar to the hippocampus and learns new incoming tasks quickly without interference from previous tasks. The STM is a collection of $n_{STM}$ small, dedicated deep generative memories (called short-term task memory -- STTM), which can each learn one unique task.

While training on an incoming task, if it is already in an STTM, the same STTM is retrained on it, otherwise a fresh STTM is allocated to the task. Additionally, if the task has been previously seen and consolidated into the LTM, then the LTM reconstructs the incoming samples for that task using the generator (hence we use a VAE), predicts labels for the reconstructions using its learner and sends these newly generated samples to the STTM allocated to this task. This provides extra samples on tasks which have been learnt previously and helps to learn them better, while also preserving the previous performance on that task to some extent.
Once all $(n_{STM})$ STTMs are exhausted, the architecture sleeps (like humans) to consolidate all tasks into the LTM and free up the STTMs for new tasks.
While asleep, the STM generates and sends samples of learnt tasks to the LTM, where these are consolidated via deep generative replay (see figure \ref{fig:dgdmn}).

While testing on task $t$ (even intermittently between tasks), if any STTM currently contains task $t$, it is used to predict the labels, else the prediction is deferred to the LTM. This allows predicting on all tasks seen uptil now (including the most recent ones) without sleeping. 
Finally note that DGR keeps track of task descriptors in dictionaries but does not use them for learning. DGDMN only uses task descriptors to recognize whether a task has been previously observed and/or the memory in which a task currently resides. This can be relaxed by using the reconstruction error from generators as a proxy for recognition (see appendix~A). Hence DGDMN still works in the absence of task descriptors.

%% file: src/exp.tex
\section{Experiments}
\label{sec:exp}

We perform experiments to demonstrate forgetting on sequential image classification tasks. We briefly describe our datasets here (details in appendix B): (a) \textbf{Permnist} is a catastrophic forgetting benchmark~\citep{kirkpatrick2017overcoming} and each task contains a fixed permutation of pixels on MNIST images, (b) \textbf{Digits} dataset involves classifying a single MNIST digit per task, (c) \textbf{TDigits} is a transformed variant of MNIST similar to Digits but with $40$ tasks for long task sequences, (d) \textbf{Shapes} contains several geometric shape classification tasks, and (e) \textbf{Hindi} contains a sequence of $8$ tasks with hindi language consonant recognition.

We compare DGDMN with several baselines for catastrophic forgetting, while choosing at least one from each category: representational overlap, learning slowdown and experience replay. These are briefly described here (implementation and hyperparameter details in appendix B): (a) \textbf{Feedforward neural networks (NN)}: To characterize forgetting in the absence of any prevention mechanism and as a reference for other approaches, (b) \textbf{Neural nets with dropout (DropNN)}: \citet{goodfellow2013empirical} suggested using dropout as a means to prevent representational overlaps and pacify catastrophic forgetting, (c) \textbf{Pseudopattern Rehearsal (PPR)}: A non-generative approach to experience replay~\citep{robins2004sequential}, (d) \textbf{Elastic Weight Consolidation (EWC)}: \citet{kirkpatrick2017overcoming} proposed using the Fisher Information Matrix for task-specific learning slowdown of weights in a neural network, and (e) \textbf{Deep Generative Replay (DGR)}: We train only the LTM from DGDMN to separate the effects of deep generative replay and dual memory architecture. This is partly similar to \citet{shin2017continual}.

In our preliminary experiments, we observed that large overparameterized networks can more easily adapt to sequentially incoming tasks, thereby partly mitigating catastrophic forgetting. So we have chosen network architectures which have to share all their parameters appropriately amongst the various tasks in a dataset to achieve reasonable \emph{joint} accuracy. This allows us to evaluate algorithms carefully while ignoring the benefits provided by overparameterization.

\subsection{Accuracy and Forgetting curves}

\begin{figure*}[tb]
\vskip 0.2in
\begin{center}
    \begin{subfigure}[th]{0.32\textwidth}
        \includegraphics[width=\textwidth]{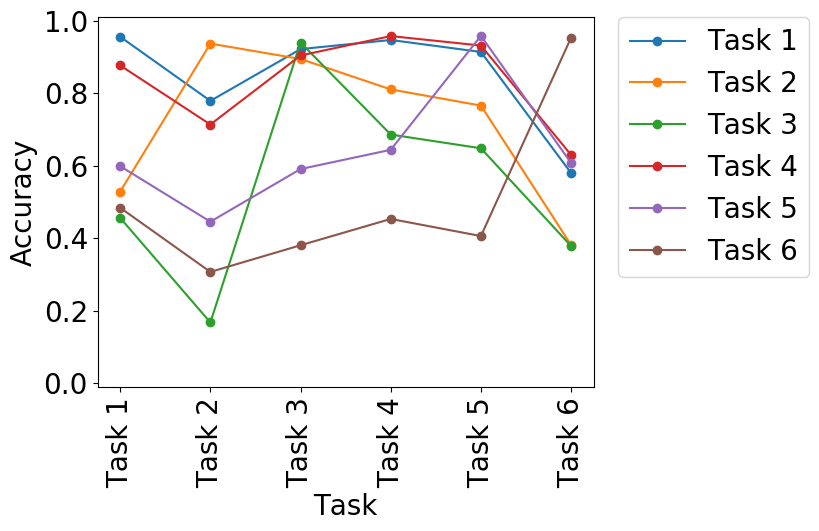}
        \caption{NN}
        \label{fig:permnistacc_NN}
    \end{subfigure}
    \begin{subfigure}[th]{0.32\textwidth}
        \includegraphics[width=\textwidth]{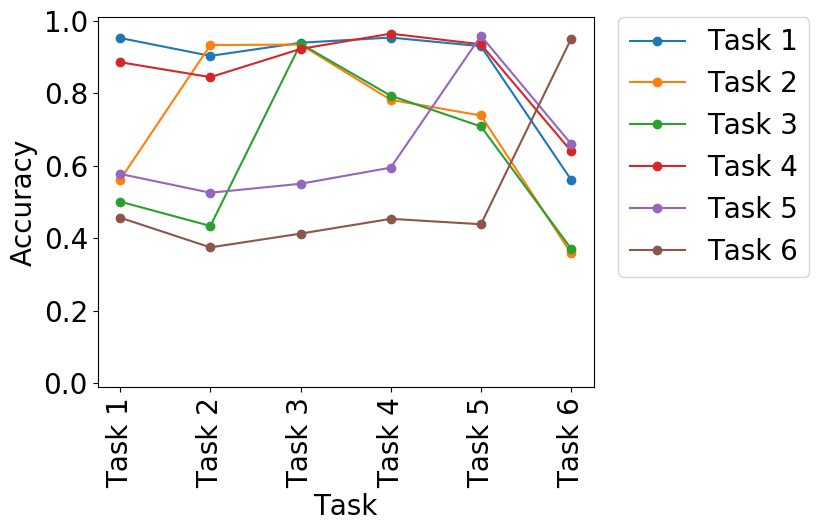}
        \caption{DropNN}
        \label{fig:permnistacc_DropNN}
    \end{subfigure}
    \begin{subfigure}[th]{0.32\textwidth}
        \includegraphics[width=\textwidth]{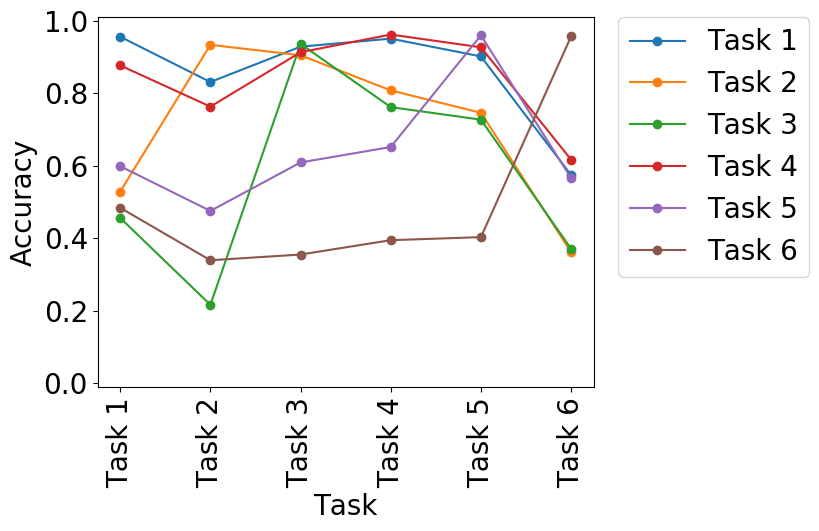}
        \caption{PPR}
        \label{fig:permnistacc_PPR}
    \end{subfigure}
    \begin{subfigure}[th]{0.32\textwidth}
        \includegraphics[width=\textwidth]{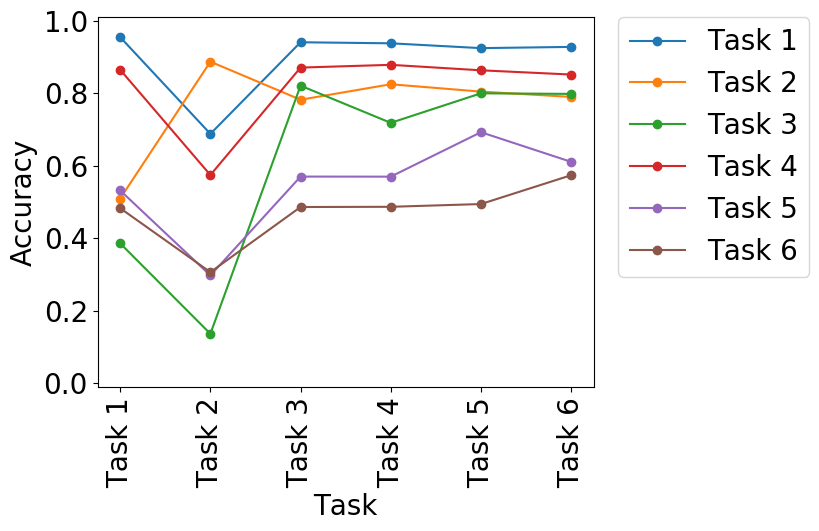}
        \caption{EWC}
        \label{fig:permnistacc_EWC}
    \end{subfigure}
    \begin{subfigure}[th]{0.32\textwidth}
        \includegraphics[width=\textwidth]{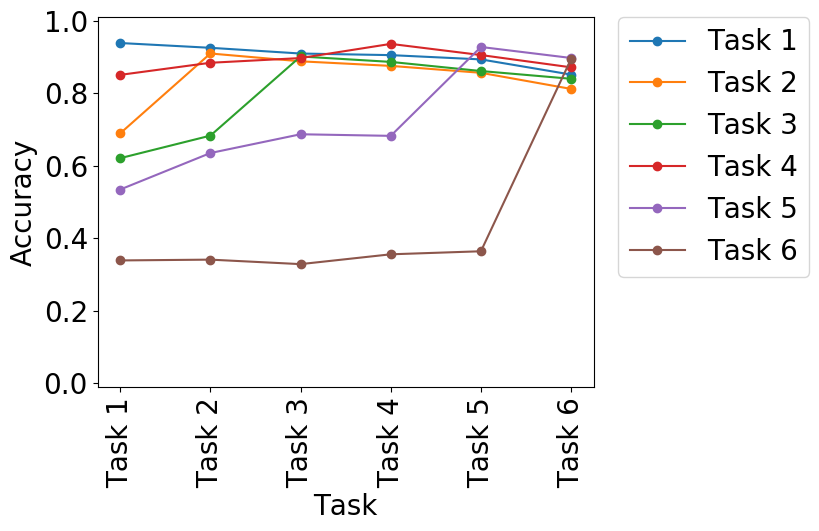}
        \caption{DGR}
        \label{fig:permnistacc_DGR}
    \end{subfigure}
    \begin{subfigure}[th]{0.32\textwidth}
        \includegraphics[width=\textwidth]{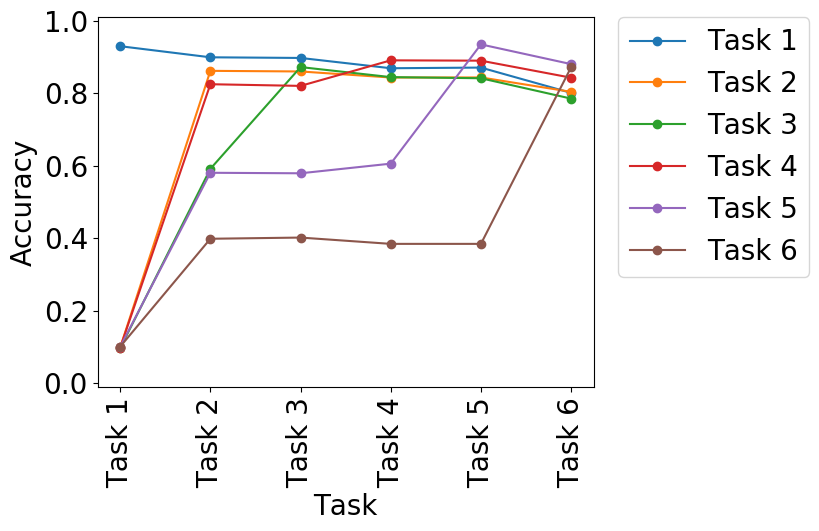}
        \caption{DGDMN}
        \label{fig:permnistacc_DGDMN}
    \end{subfigure}
\caption{Accuracy curves for Permnist (x: tasks seen, y: classification accuracy on task).}
\label{fig:permnistacc}
\end{center}
\vskip -0.2in
\end{figure*}

\begin{figure*}[bt]
\vskip 0.2in
\begin{center}
    \begin{subfigure}[th]{0.32\textwidth}
        \includegraphics[width=\textwidth]{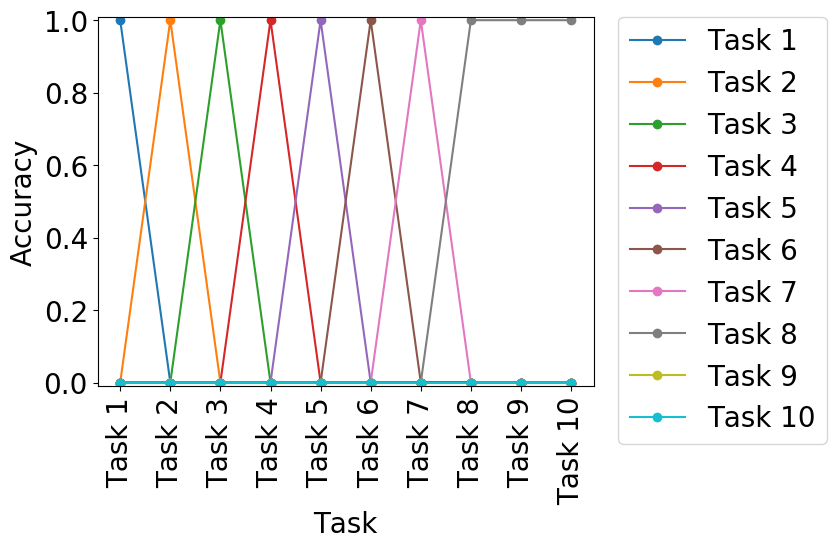}
        \caption{NN}
        \label{fig:digitsacc_NN}
    \end{subfigure}
    \begin{subfigure}[th]{0.32\textwidth}
        \includegraphics[width=\textwidth]{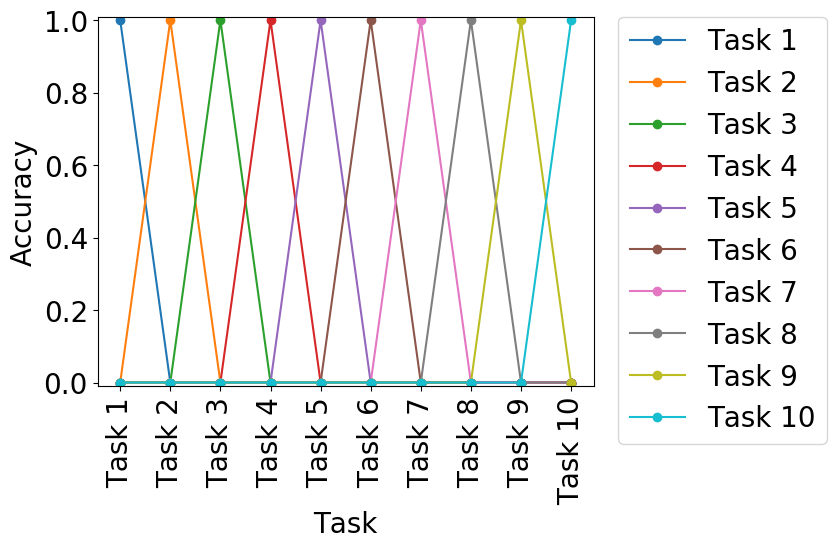}
        \caption{DropNN}
        \label{fig:digitsacc_DropNN}
    \end{subfigure}
    \begin{subfigure}[th]{0.32\textwidth}
        \includegraphics[width=\textwidth]{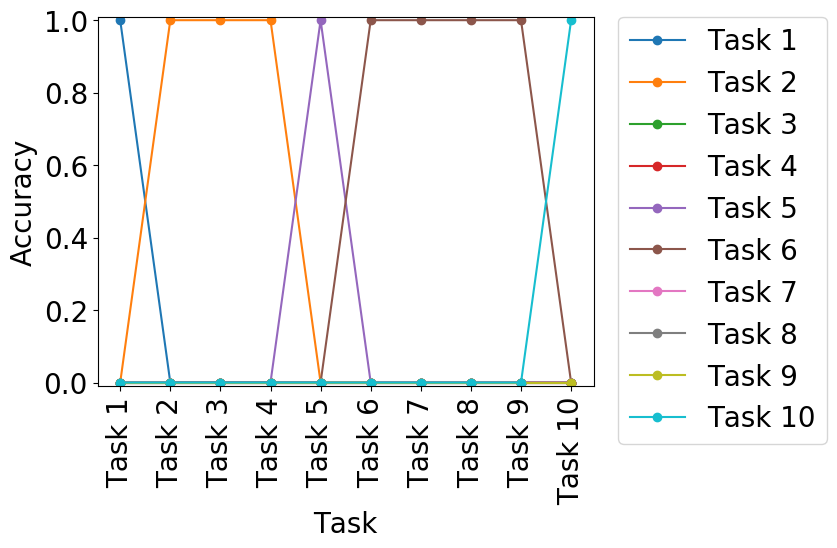}
        \caption{PPR}
        \label{fig:digitsacc_PPR}
    \end{subfigure}
    \begin{subfigure}[th]{0.32\textwidth}
        \includegraphics[width=\textwidth]{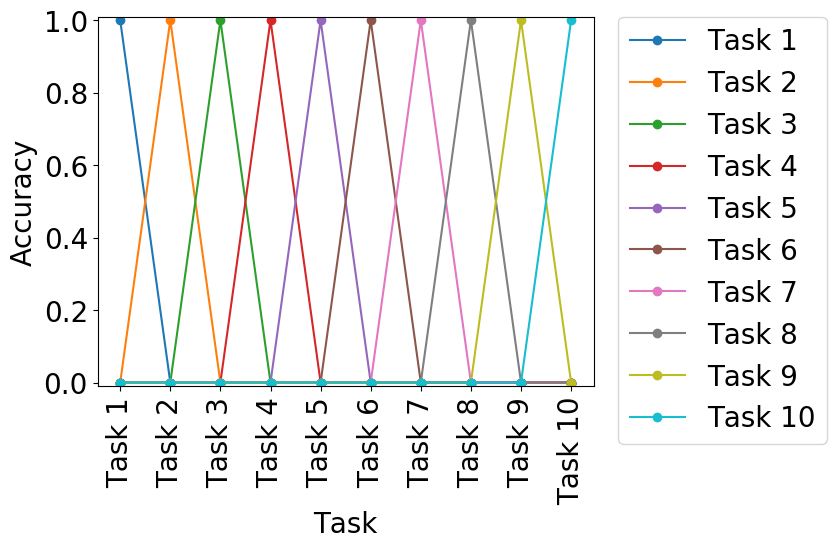}
        \caption{EWC}
        \label{fig:digitsacc_EWC}
    \end{subfigure}
    \begin{subfigure}[th]{0.32\textwidth}
        \includegraphics[width=\textwidth]{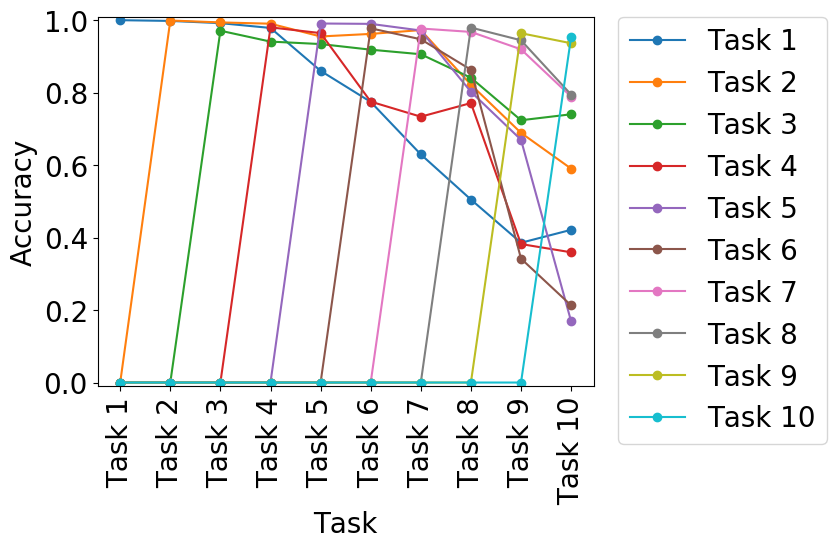}
        \caption{DGR}
        \label{fig:digitsacc_DGR}
    \end{subfigure}
    \begin{subfigure}[th]{0.32\textwidth}
        \includegraphics[width=\textwidth]{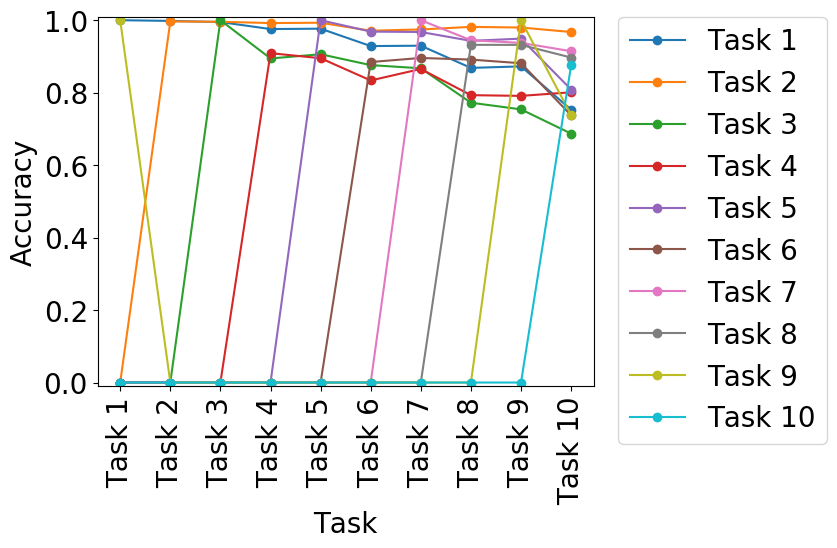}
        \caption{DGDMN}
        \label{fig:digitsacc_DGDMN}
    \end{subfigure}
\caption{Accuracy curves for Digits (x: tasks seen, y: classification accuracy on task).}
\label{fig:digitsacc}
\end{center}
\vskip -0.2in
\end{figure*}

We trained DGDMN and all baselines sequentially on the image classification tasks of Permnist, Digits, Shapes and Hindi datasets (separately). Due to space constraints, we show results on the Shapes and Hindi datasets in appendix A. The classification accuracy on a held out test set for each task, after training on the $t^{th}$ task has been shown in figures \ref{fig:permnistacc} and \ref{fig:digitsacc}. We used the same network architecture for NN, PPR, EWC, learner in DGR, and learner in the LTM of DGDMN for a given dataset. DropNN had intermediate dropouts after hidden layers (details in appendix B).

We observe from figures \ref{fig:permnistacc} and \ref{fig:digitsacc}, that NN and DropNN forget catastrophically while learning and perform similarly. We verified the same on other datasets in Appendix A.
EWC performs better than NN and DropNN, but rapidly slows down learning on many weights and effectively stagnates after Task 3 (e.g.\ see Tasks 5 and 6 in figure \ref{fig:permnistacc_EWC}). The learning slowdown on weights hinders EWC from reusing those weights later to jointly discover common structures between tasks. Note that the networks do have the capacity to learn all tasks and our generative replay based algorithms DGR and DGDMN indeed learn all tasks sequentially with the same learner networks.

\begin{figure*}[th]
\vskip 0.2in
\begin{center}
    \begin{subfigure}[th]{0.4\textwidth}
        \includegraphics[width=\textwidth]{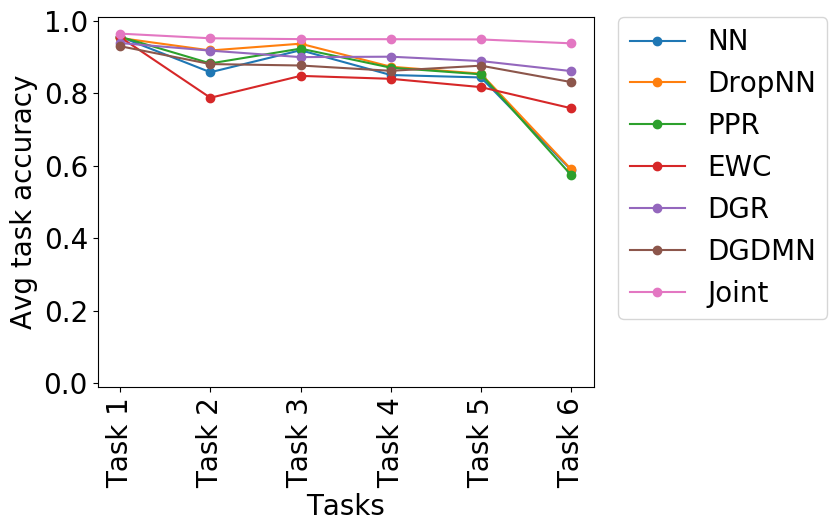}
        \caption{Permnist}
        \label{fig:permnist_avgforget}
    \end{subfigure}
    \begin{subfigure}[th]{0.4\textwidth}
        \includegraphics[width=\textwidth]{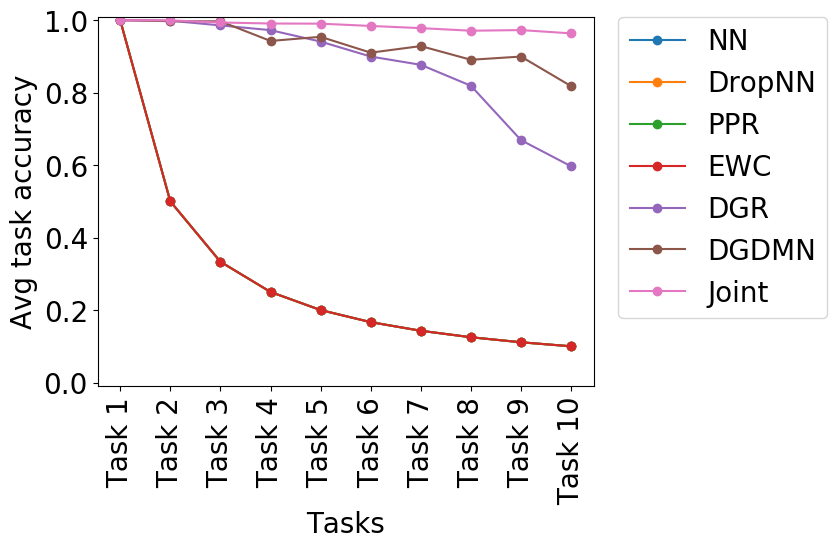}
        \caption{Digits}
        \label{fig:digits_avgforget}
    \end{subfigure}
\caption{Forgetting curves (x: tasks seen, y: avg classification accuracy on tasks seen).}
\label{fig:avgforget}
\end{center}
\vskip -0.2in
\end{figure*}

Further, we observed heavy forgetting on Digits (figure \ref{fig:digitsacc}) for most baselines, which is expected because all samples in the $t^{th}$ task have a single label ($t$) and the $t^{th}$ task can be learnt on its own by setting the $t^{th}$ bias of the final softmax layer to be high and the other biases to be low. Such sequential tasks cause networks to forget catastrophically. We observed that NN, DropNN, PPR and EWC learnt only the task being trained on and forgot all previous knowledge immediately. Sometimes, we also observed saturation due to the softmax bias being set very high and then being unable to recover from it. PPR showed severe saturation since its replay prevented it from coming out of the saturation.

DGR and DGDMN still retain performance on all tasks of Digits, since they replay generated samples from previous tasks. The average forgetting on all tasks $\in \{1, \hdots, t\}$, after training on the $t^{th}$ task (for both Digits and Permnist) is shown in figure \ref{fig:avgforget}. For absolute reference, the accuracy of NN by training it jointly on all tasks uptil the $t^{th}$ task has also been shown for each $t$. This also shows that DGR and DGDMN consistently outperform baselines in terms of retained average accuracy. In figure \ref{fig:digits_avgforget}, NN, DropNN, PPR and EWC follow nearly overlapping curves ($acc \approx \frac{1}{t}$) since they are only able to learn one task at a time.
Though PPR also involves experience replay, it is not able to preserve its learnt mapping by randomly sampling points from its domain and hence forgets catastrophically. These observations substantiate our claim that a replay mechanism must be generative and model the input distribution accurately. We observed similar results on other datasets (appendix A).

\begin{table}[h]
\caption{Average accuracies for all algorithms.}
\label{tab:acc}
\vskip -0.1in
\begin{center}
\begin{small}
\begin{sc}
\begin{tabular}{l|cccr}
\toprule
Algorithm & Digits & Permnist & Shapes & Hindi \\
\midrule
NN & 0.1 & 0.588 & 0.167 & 0.125 \\
DropNN & 0.1 & 0.59 & 0.167 & 0.125 \\
PPR & 0.1 & 0.574 & 0.167 & 0.134 \\
EWC & 0.1 & 0.758 & 0.167 & 0.125 \\
DGR & 0.596 & \textbf{0.861} & 0.661 & \textbf{0.731} \\
DGDMN & \textbf{0.818} & 0.831 & \textbf{0.722} & 0.658 \\
\bottomrule
\end{tabular}
\end{sc}
\end{small}
\end{center}
\vskip -0.1in
\end{table}

\begin{table}[h]
\caption{Backward transfer for all algorithms.}
\label{tab:bwt}
\vskip -0.1in
\begin{center}
\begin{small}
\begin{sc}
\begin{tabular}{l|cccr}
\toprule
Algorithm & Digits & Permnist & Shapes & Hindi \\
\midrule
NN & -0.778 & -0.434 & -0.4 & -1.0 \\
DropNN & -1.0 & -0.43 & -0.8 & -1.0 \\
PPR & -0.444 & -0.452 &-0.2 & -0.989 \\
EWC & -1.0 & -0.05 & -1.0 & -1.0 \\
DGR & -0.425 & \textbf{-0.068} & -0.288 & \textbf{-0.270} \\
DGDMN & \textbf{-0.15} & -0.075 & \textbf{-0.261} & -0.335 \\
\bottomrule
\end{tabular}
\end{sc}
\end{small}
\end{center}
\vskip -0.2in
\end{table}

We show the final average accuracies (ACC) and backward transfer (BWT) between tasks in tables \ref{tab:acc} and \ref{tab:bwt} respectively. NN, DropNN, PPR and EWC get near random accuracies on all datasets except Permnist due to catastrophic forgetting. DGDMN and DGR perform similarly and outperform other baselines on ACC while having the least negative BWT. Since backward transfer is a direct measure of forgetting, this also shows that we effectively mitigate catastrophic forgetting and avoid inter-task interference.
We point out that datasets like Digits should be considered important benchmarks for continual learning since they have low correlation between samples of different tasks and promote overfitting to the new incoming task thereby causing catastrophic forgetting. Being able to retain performance on such task sequences is a strong indicator of the effectiveness of a continual learning algorithm.

\subsection{Connections to complementary learning systems and sleep}
\label{subsec:longseq}

To differentiate between DGDMN and DGR, we trained both of them on a long sequence of $40$ tasks from TDigits dataset. We limited $N_{max}$ to $120,000$ samples for this task to explore the case where the LTM in DGDMN (DGM in DGR) cannot regenerate many samples and has to forget some tasks. At least $\kappa=0.05$ fraction of memory was ensured for new task samples and consolidation in DGDMN happened after $n_{STM}=5$ tasks.

\begin{figure*}[th]
\vskip 0.2in
\begin{center}
    \begin{subfigure}[th]{0.32\textwidth}
        \centerline{\includegraphics[width=\textwidth]{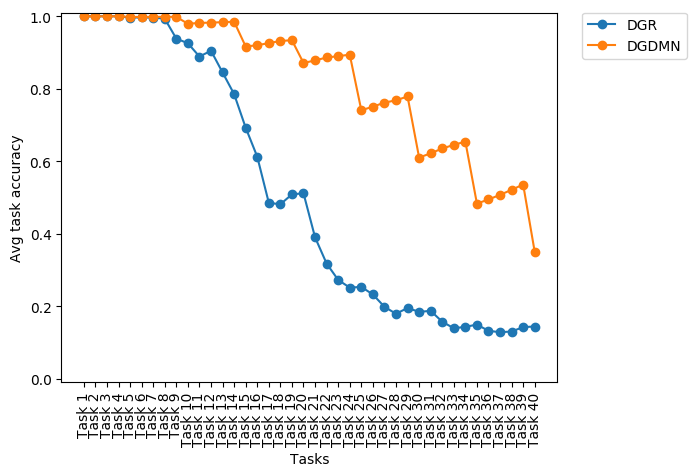}}
        \caption{}
        \label{fig:tdigits_avgforget}
    \end{subfigure}
    \begin{subfigure}[th]{0.32\textwidth}
        \centerline{\includegraphics[width=\textwidth]{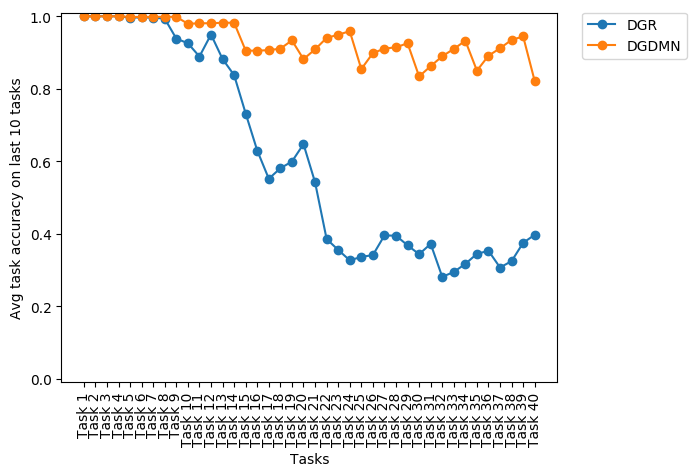}}
        \caption{}
        \label{fig:tdigits_avgforget_10}
    \end{subfigure}
    \begin{subfigure}[th]{0.32\textwidth}
        \includegraphics[width=\textwidth]{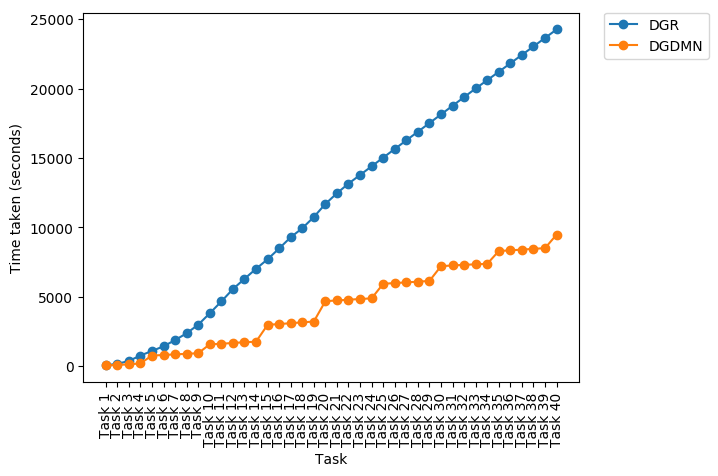}
        \caption{}
        \label{fig:tdigits_time}
    \end{subfigure}
\caption{Accuracy and training time for DGDMN and DGR on TDigits: (a) Accuracy on tasks seen so far, (b) Accuracy on last 10 tasks seen, (c) Training time}
\label{fig:tdigits_avgforget_curves}
\end{center}
\vskip -0.2in
\end{figure*}

The average forgetting curves are plotted in figure \ref{fig:tdigits_avgforget} and show that forgetting is gradual and not catastrophic. DGDMN retains more accuracy on all tasks as compared to DGR and is faster to train as shown by figure \ref{fig:tdigits_time}.
This is because DGR consolidates its DGM after every task. Since LTM is a large memory and requires more samples to consolidate, it trains slower. Further, the DGM's self-generated slightly erroneous samples compound errors quite fast. On the other hand, DGDMN uses small STTMs to learn single tasks faster and with low error. Consequently, the LTM consolidates less often and sees more accurate samples, hence its error accumulates much slower.
Lastly, DGDMN stays around $90\%$ average accuracy on the most recently observed $10$ tasks (figure \ref{fig:tdigits_avgforget_10}), whereas DGR propagates errors too fast and also fails on this metric eventually.

Dual memory architecture and periodic sleep has emerged naturally in humans as a scalable design choice. Though sleeping is a dangerous behavior for any organism due to risk of being attacked by a predator, it has still survived eons of evolution~\citep{joiner2016unraveling} and most organisms with even a slightly developed nervous system (centralized or diffuse) still exhibit either sleep or light-resting behavior~\citep{nath2017jellyfish}. This experiment partly sheds light on the importance of dual memory architecture intertwined with periodic sleep, without which learning would be highly time consuming and short lived (as in DGR).

%% file: src/anal.tex
\section{Analysis and discussion}
\label{sec:anal}

We next show that DGDMN shares some remarkable characteristics with the human memory and present a discussion of some relevant ideas. Due to space constraints, we have deferred some visualizations of the learnt latent structures to appendix A. The hyperparameters of DGDMN ($\kappa$ and $n_{STM}$) admit intuitive interpretations and can be tuned with simple heuristics (see appendix B).

\begin{figure*}[th]
\vskip 0.2in
\begin{center}
    \begin{subfigure}[b]{0.2\textwidth}
        \includegraphics[width=\textwidth]{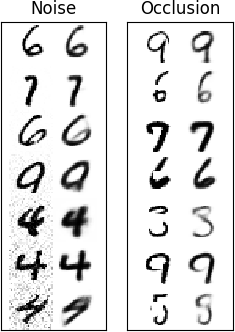}
        \caption{}
        \label{fig:recon}
    \end{subfigure}
    \begin{subfigure}[b]{0.32\textwidth}
        \includegraphics[width=\textwidth]{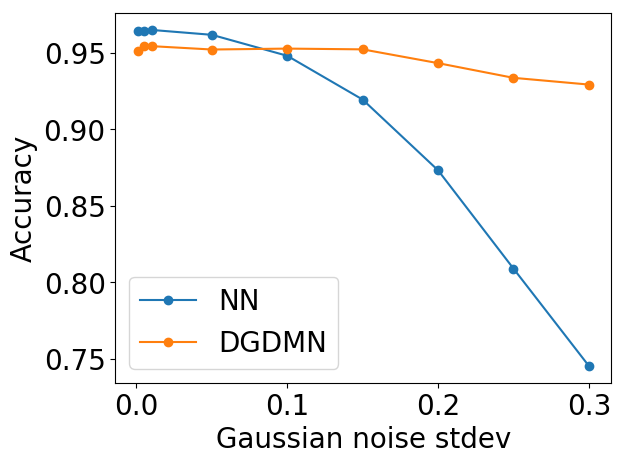}
        \caption{}
        \label{fig:noisy}
    \end{subfigure}
    \begin{subfigure}[b]{0.32\textwidth}
        \includegraphics[width=\textwidth]{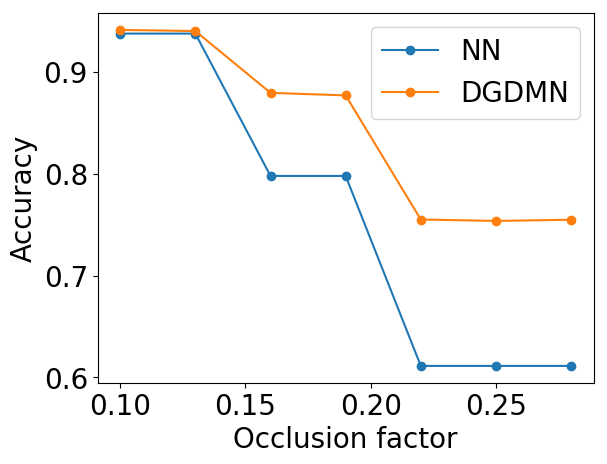}
        \caption{}
        \label{fig:occluded}
    \end{subfigure}
\caption{LTM is robust to noisy and occluded images and exhibits smoother degradation in classification accuracy because of its denoising reconstructive properties: (a) LTM reconstruction from noisy and occluded digits, (b) Classification accuracy with increasing gaussian noise, and (c) Classification accuracy with increasing occlusion factor.}
\label{fig:resilience}
\end{center}
\vskip -0.2in
\end{figure*}

\textbf{Resilience to noise and occlusion}:\label{subsec:resilience} We have used a VAE to be able to reconstruct all samples, which helps to recognize task examples (appendix A) and also makes our model resilient to noise, distortion and occlusion. 
We tested our LTM model and a NN model by jointly training on uncorrupted Digits data and testing on noisy and occluded images. Figure \ref{fig:resilience} shows that the LTM is more robust to noise and occlusion due to its denoising reconstructive properties.

\textbf{The choice of underlying generative model}:\label{subsec:choicegen} Our architecture is agnostic to the choice of the underlying generative model as long as the generator can generate reliable samples and reconstruct incoming samples accurately. Hence, apart from VAEs, variants of Generative Adversarial Networks like BiGANs~\citep{donahue2017adversarial}, ALI~\citep{dumoulin2017adversarially} and AVB~\citep{mescheder2017adversarial} can be used depending on the modeled domain.

\textbf{Connections to knowledge distillation}:\label{sec:distill} Previous works on (joint) multitask learning have also proposed approaches to learn individual tasks with small networks and then ``distilling'' them jointly into a larger network~\citep{rusu2015policy}. Such distillation can sometimes improve performance on individual tasks if they share structure and at other times mitigate inter-task interference due to refinement of learnt functions while distilling~\citep{parisotto2016actor}. Similarly, due to refinement and compression during consolidation phase, DGDMN is also able to learn joint task structure effectively while mitigating interference between tasks.

\textbf{Learning from streaming data}: We have presently formulated our setup with task descriptors to compare it with existing approaches in the continual learning literature, but we emphasize that having no dependence on task descriptors is an essential step to learn continually from streaming data. Our approach allows online recognition of task samples via a reconstructive generative model and is applicable in domains with directly streaming data without any task descriptors unlike most previous approaches which make explicit use of task descriptors~\cite{zenke2017continual,kirkpatrick2017overcoming,rebuffi2017icarl,lopez2017gradient} (see appendix A). This would allow DGDMN to be used for learning policies over many tasks via reinforcement learning without explicit replay memories, and we plan to explore this in future work.

\textbf{Approaches based on synaptic consolidation}:\label{subsec:synconsol} Though our architecture draws inspiration from complementary learning systems and experience replay in the human brain, there is also neuroscientific evidence for synaptic consolidation in the human brain like in \citep{kirkpatrick2017overcoming} and \citep{zenke2017continual}. It might be interesting to explore how synaptic consolidation can be incorporated in our dual memory architecture without causing stagnation and we leave this to future work.

%% file: src/conc.tex
\section{Conclusion}
\label{sec:conc}

In this work, we have developed a continual learning architecture to avert catastrophic forgetting. 
Our dual memory architecture emulates the complementary learning systems in the human brain and maintains a consolidated long-term memory via generative replay of past experiences.
We have shown that generative replay performs the best for long-term performance retention and scales well along with a dual memory architecture via our experiments. Moreover, our architecture displays significant parallels with the human memory system and provides useful insights about the connection between sleep and learning in humans.

%% file: src/appendixA.tex
\section{Appendix A}
\label{sec:appendixA}

\subsection{Repeated tasks and revision}
\label{subsec:revision}

\begin{figure*}[th]
\vskip 0.2in
\begin{center}
    \begin{subfigure}[th]{0.32\textwidth}
        \centerline{\includegraphics[width=\textwidth]{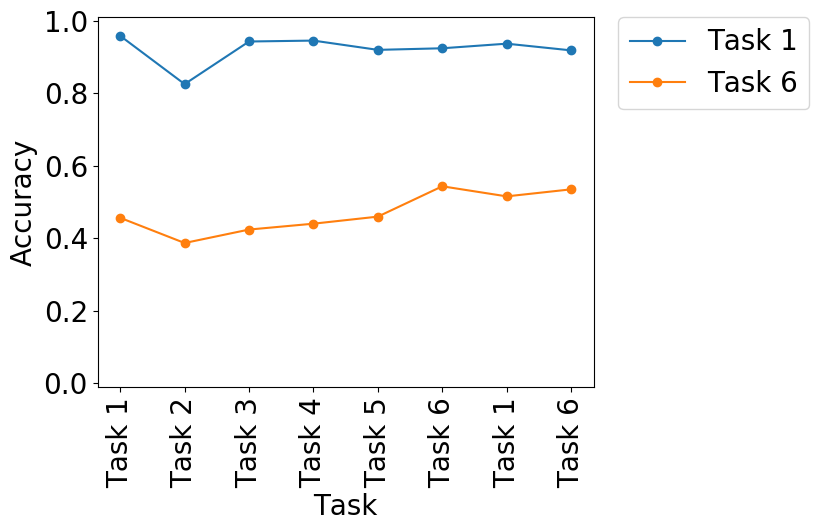}}
        \caption{}
        \label{fig:revision_EWC}
    \end{subfigure}
    \begin{subfigure}[th]{0.32\textwidth}
        \centerline{\includegraphics[width=\textwidth]{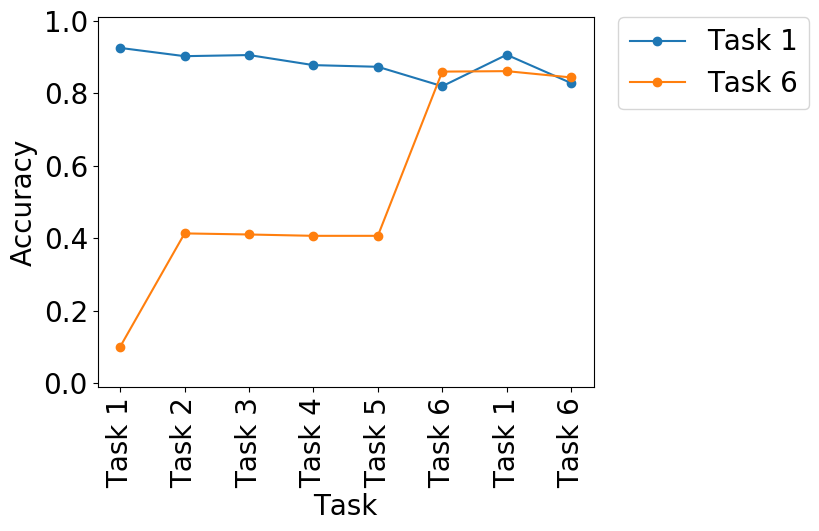}}
        \caption{}
        \label{fig:revision_DGDMN}
    \end{subfigure}
\caption{Accuracy curves when tasks are revised: (a) EWC, (b) GEM, and (c) DGDMN.}
\label{fig:revision}
\end{center}
\vskip -0.2in
\end{figure*}

It is well known in psychology literature that human learning improves via revision~\citep{kahana2005spacing, cepeda2006distributed}. We show performance of EWC and DGDMN on Permnist, when some tasks are repeated (figure \ref{fig:revision}). DGR performs very similar to DGDMN, hence we omit it. EWC stagnates and once learning has slowed down on the weights important for Task 1, the weights cannot be changed again, not even for improving Task 1. Further, it did not learn Task 6 the first time and revision does not help either. However, DGDMN learns all tasks uptil Task 6 and then improves by revising Task 1 and 6 again.
We point out that methods involving freezing (or slowdown) of learning often do not learn well via revision since they do not have any means of identifying tasks and unfreezing the previously frozen weights when the task is re-encountered. While many previous works do not investigate revision, it is crucial for learning continuously and should improve performance on tasks. The ability to learn from correlated task samples and revision makes our architecture functionally similar to that of humans.

\subsection{Experiments on other datasets}

\begin{figure*}[tb]
\vskip 0.2in
\begin{center}
    \begin{subfigure}[th]{0.32\textwidth}
        \includegraphics[width=\textwidth]{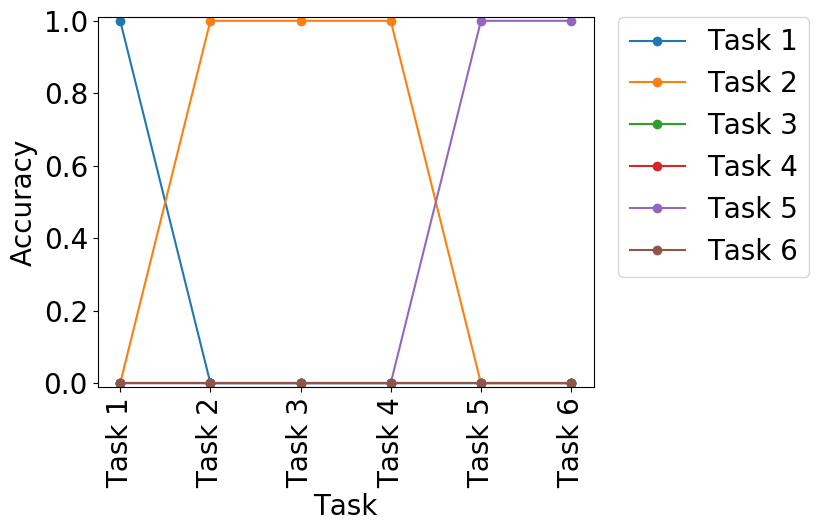}
        \caption{CNN}
        \label{fig:shapesacc_CNN}
    \end{subfigure}
    \begin{subfigure}[th]{0.32\textwidth}
        \includegraphics[width=\textwidth]{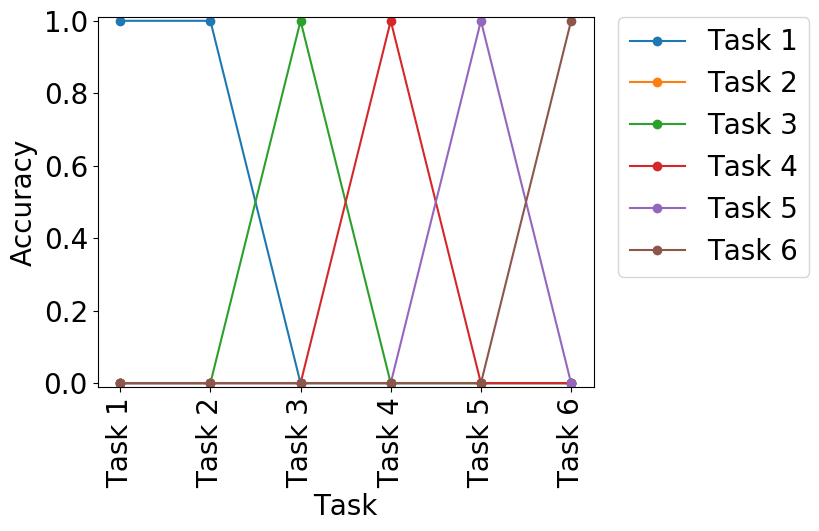}
        \caption{DropCNN}
        \label{fig:shapesacc_DropCNN}
    \end{subfigure}
    \begin{subfigure}[th]{0.32\textwidth}
        \includegraphics[width=\textwidth]{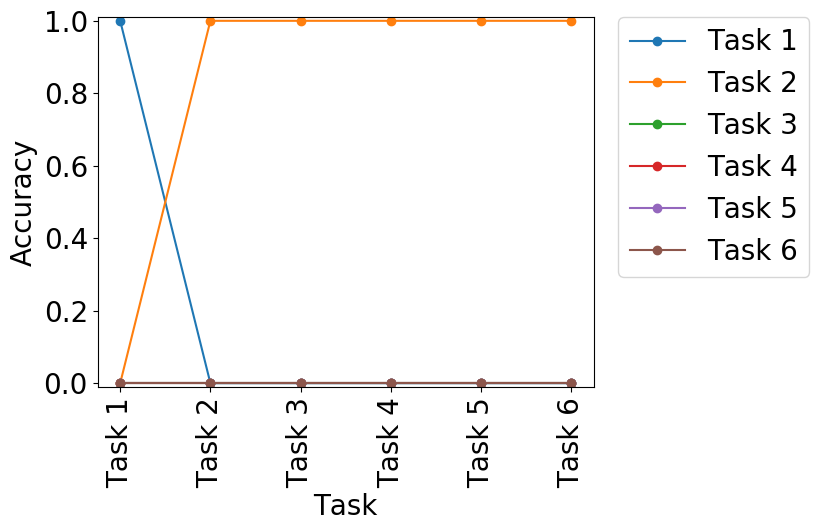}
        \caption{PPR}
        \label{fig:shapesacc_PPR}
    \end{subfigure}
    \begin{subfigure}[th]{0.32\textwidth}
        \includegraphics[width=\textwidth]{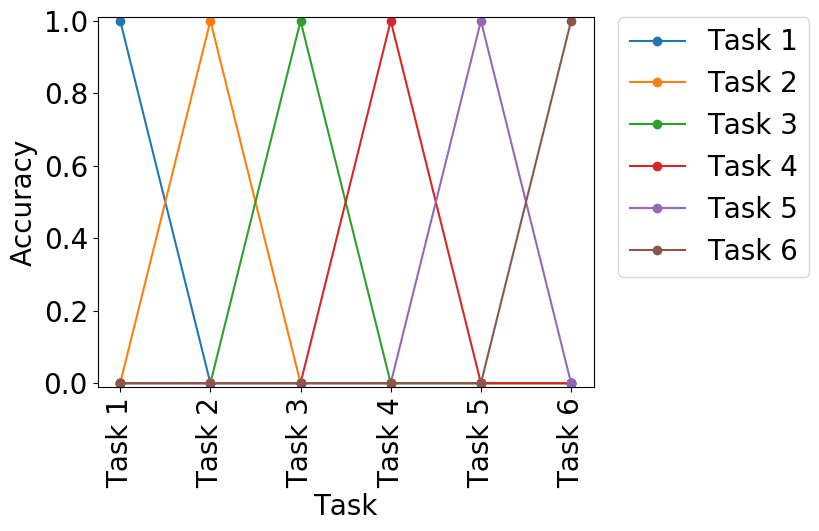}
        \caption{EWC}
        \label{fig:shapesacc_EWC}
    \end{subfigure}
    \begin{subfigure}[th]{0.32\textwidth}
        \includegraphics[width=\textwidth]{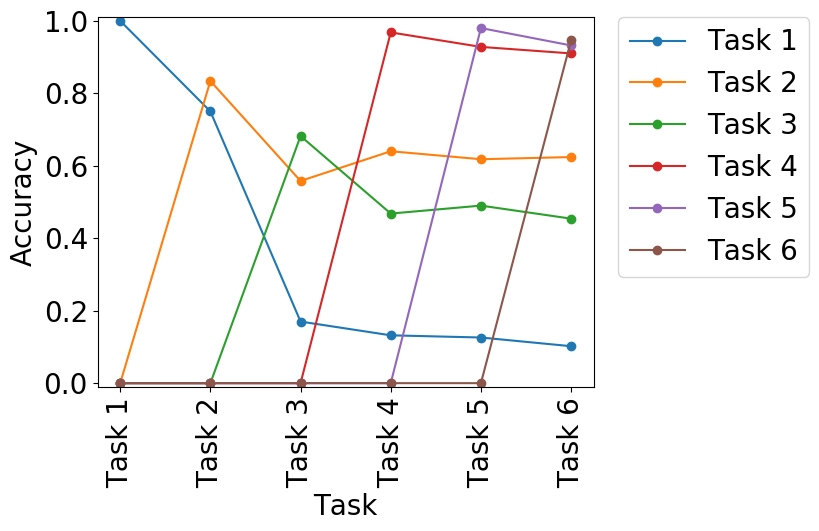}
        \caption{DGR}
        \label{fig:shapesacc_DGR}
    \end{subfigure}
    \begin{subfigure}[th]{0.32\textwidth}
        \includegraphics[width=\textwidth]{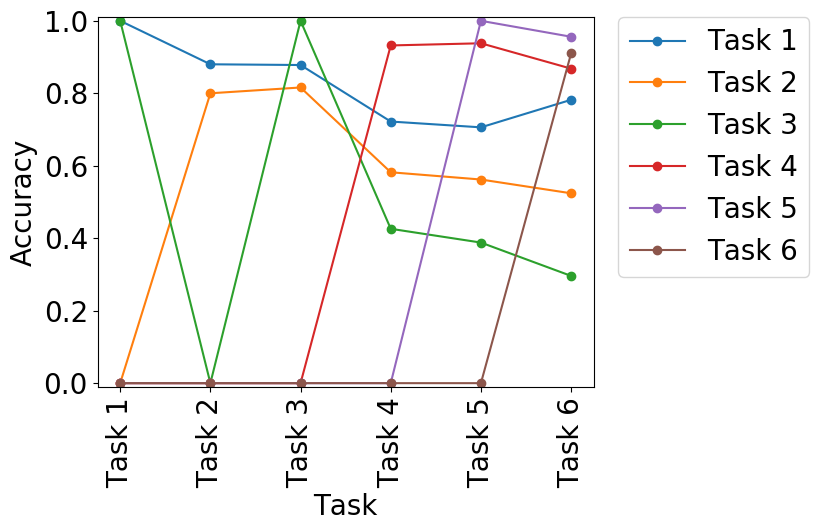}
        \caption{DGDMN}
        \label{fig:shapesacc_DGDMN}
    \end{subfigure}
\caption{Accuracy curves for Shapes (x: tasks seen, y: classification accuracy on task).}
\label{fig:shapesacc}
\end{center}
\vskip -0.2in
\end{figure*}

\begin{figure*}[bt]
\vskip 0.2in
\begin{center}
    \begin{subfigure}[th]{0.32\textwidth}
        \includegraphics[width=\textwidth]{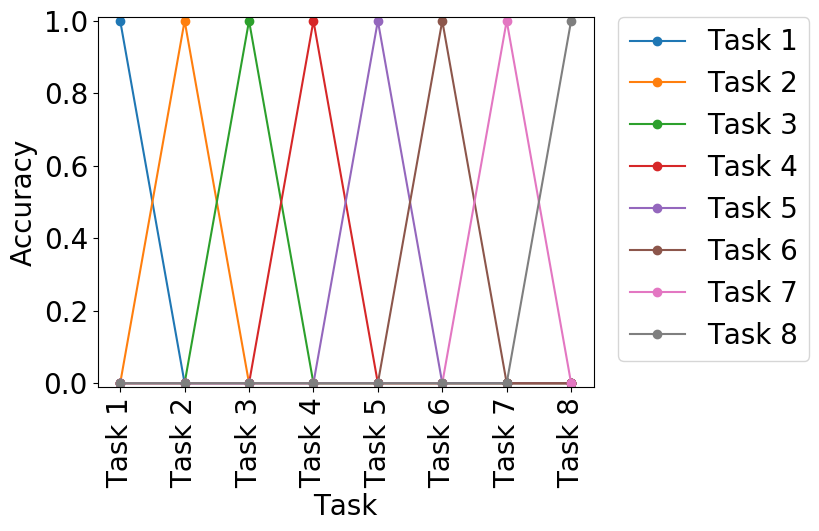}
        \caption{CNN}
        \label{fig:hindiacc_CNN}
    \end{subfigure}
    \begin{subfigure}[th]{0.32\textwidth}
        \includegraphics[width=\textwidth]{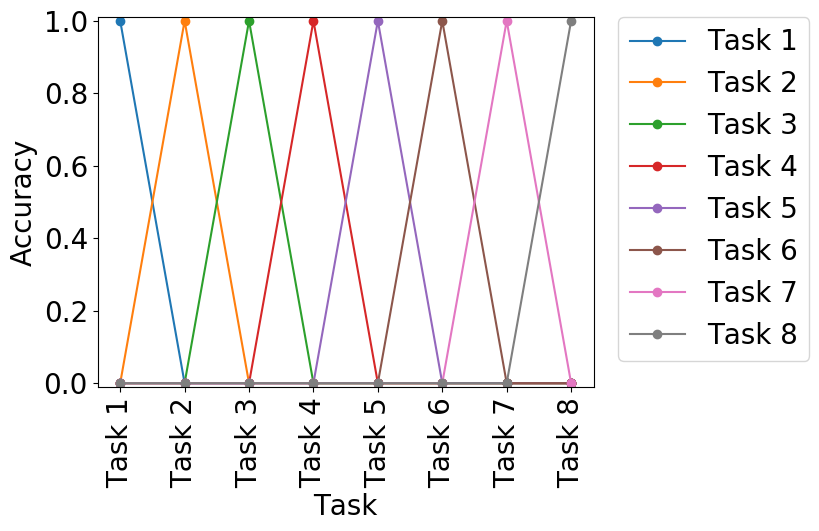}
        \caption{DropCNN}
        \label{fig:hindiacc_DropCNN}
    \end{subfigure}
    \begin{subfigure}[th]{0.32\textwidth}
        \includegraphics[width=\textwidth]{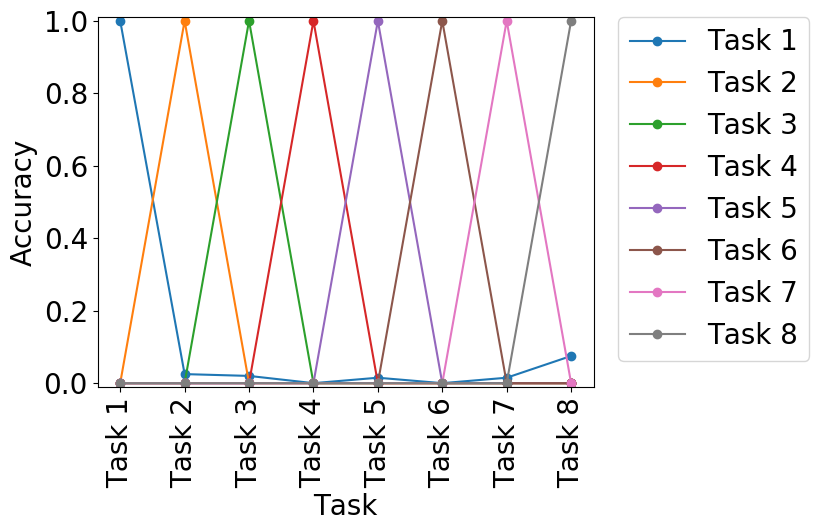}
        \caption{PPR}
        \label{fig:hindiacc_PPR}
    \end{subfigure}
    \begin{subfigure}[th]{0.32\textwidth}
        \includegraphics[width=\textwidth]{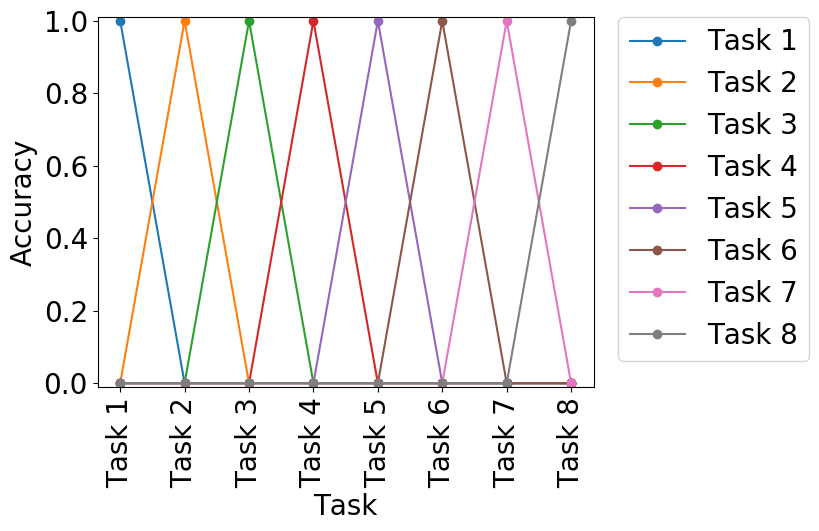}
        \caption{EWC}
        \label{fig:hindiacc_EWC}
    \end{subfigure}
    \begin{subfigure}[th]{0.32\textwidth}
        \includegraphics[width=\textwidth]{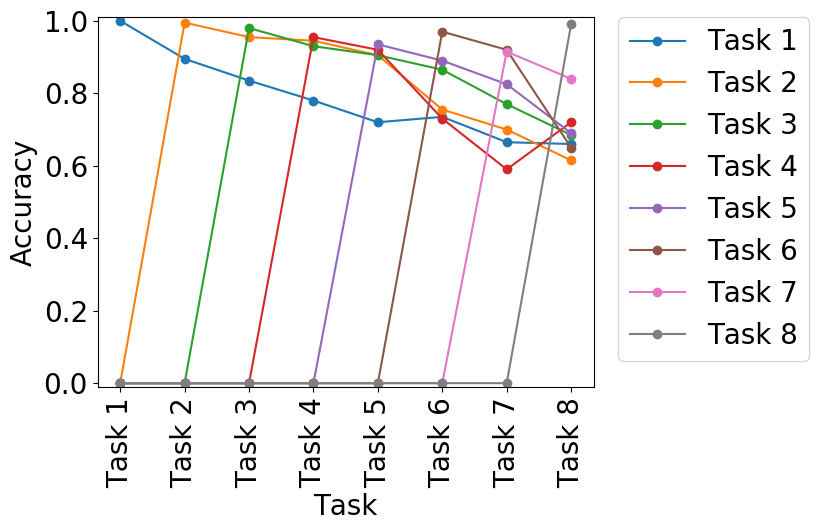}
        \caption{DGR}
        \label{fig:hindiacc_DGR}
    \end{subfigure}
    \begin{subfigure}[th]{0.32\textwidth}
        \includegraphics[width=\textwidth]{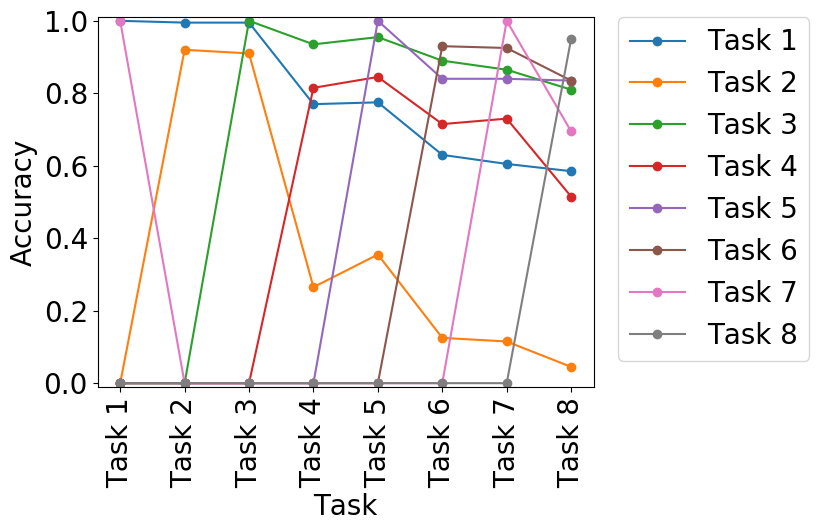}
        \caption{DGDMN}
        \label{fig:hindiacc_DGDMN}
    \end{subfigure}
\caption{Accuracy curves for Hindi (x: tasks seen, y: classification accuracy on task).}
\label{fig:hindiacc}
\end{center}
\vskip -0.2in
\end{figure*}

\begin{figure}[h]
\vskip 0.2in
\begin{center}
    \begin{subfigure}[th]{0.35\textwidth}
        \includegraphics[width=\textwidth]{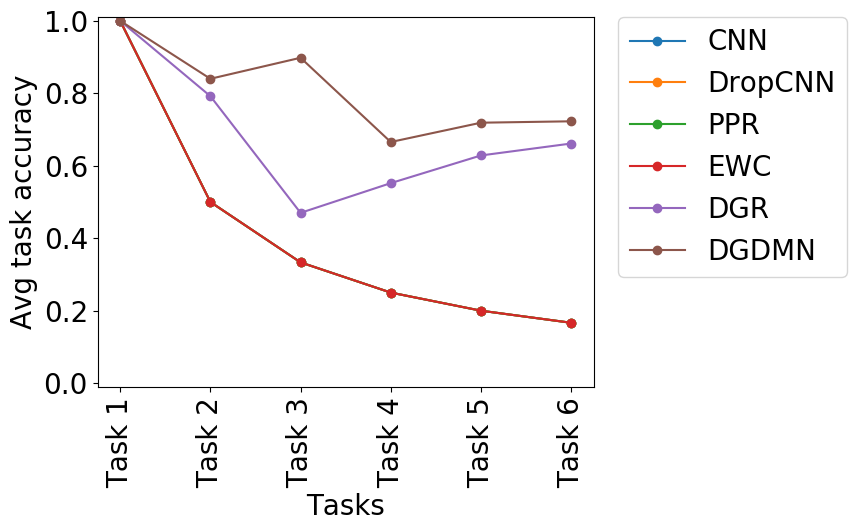}
        \caption{Shapes}
        \label{fig:shapes_avgforget}
    \end{subfigure}
    \begin{subfigure}[th]{0.35\textwidth}
        \includegraphics[width=\textwidth]{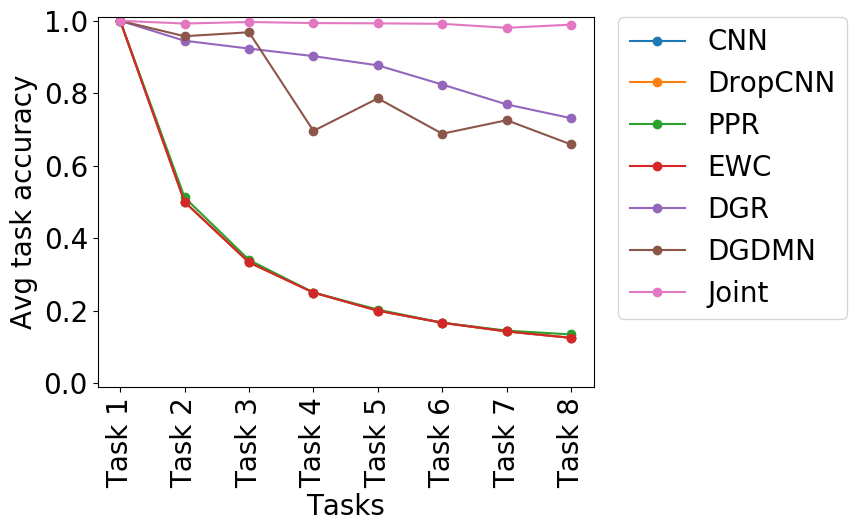}
        \caption{Hindi}
        \label{fig:hindi_avgforget}
    \end{subfigure}
\caption{Forgetting curves on Shapes and Hindi dataset (x: tasks seen, y: avg classification accuracy on tasks seen).}
\label{fig:avgforgetsupp}
\end{center}
\vskip -0.2in
\end{figure}

In this section, we present more experiments on the Shapes and the Hindi dataset, which contain sequences of tasks with geometric shapes and hindi consonants recognition respectively. We observed similar forgetting patterns as on the Digits dataset in section \ref{sec:exp}. All baselines exhibited catastrophic forgetting on these sequences of tasks, but DGR and DGDMN were able to learn the task structure sequentially (figures \ref{fig:shapesacc}, \ref{fig:hindiacc}). The same is reflected in the average forgetting curves in figure \ref{fig:avgforgetsupp}.

\subsection{Jointly vs. sequentially learnt structure}

\begin{figure}[h]
\vskip 0.2in
\begin{center}
    \begin{subfigure}[b]{0.3\textwidth}
        \includegraphics[width=\textwidth]{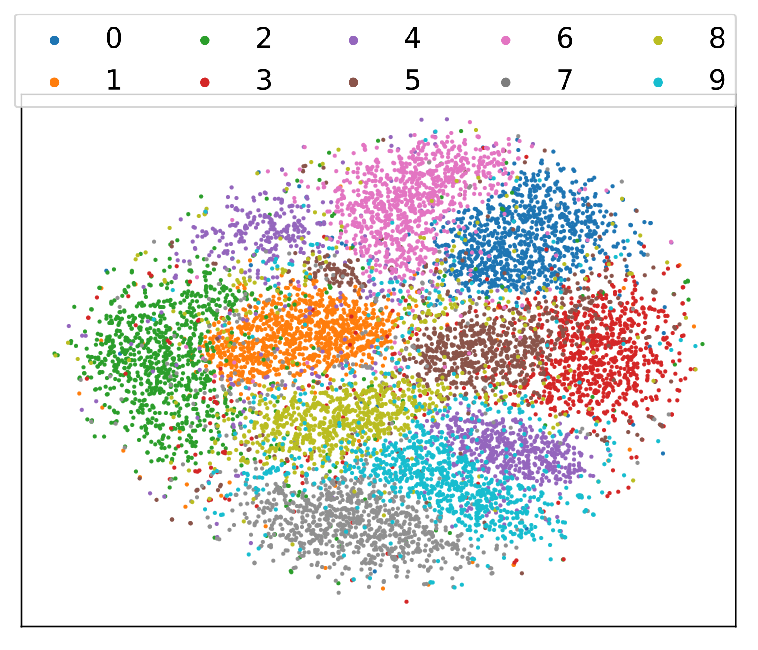}
        \caption{}
        \label{fig:tsneJ}
    \end{subfigure}
    \begin{subfigure}[b]{0.35\textwidth}
        \includegraphics[width=\textwidth]{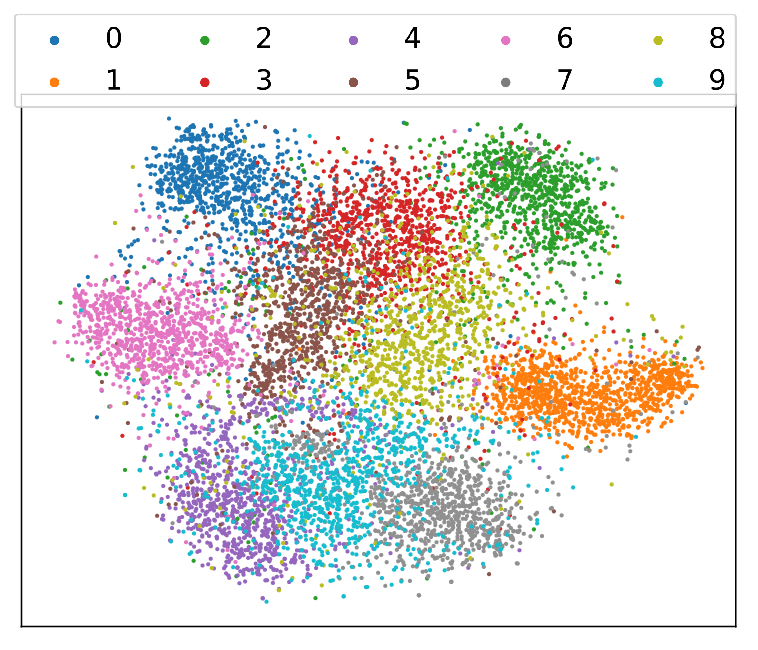}
        \caption{}
        \label{fig:tsneS}
    \end{subfigure}
\caption{t-SNE embedding for latent vectors of the VAE generator on Digits dataset when: (a) tasks are learnt jointly, and (b) tasks are learnt sequentially.}
\label{fig:tsne}
\end{center}
\vskip -0.2in
\end{figure}

To explore whether learning tasks sequentially results in a similar structure as learning them jointly, we visualized t-SNE~\citep{maaten2008visualizing} embeddings of the latent vectors of the LTM generator (VAE) in DGDMN after training it: (a) jointly over all tasks (Figure \ref{fig:tsneJ}), and (b) sequentially over tasks seen one at a time (Figure \ref{fig:tsneS}) on the Digits dataset. To maintain consistency, we used the same random seed in t-SNE for both joint and sequential embeddings.

We observe that the LTM's latent space effectively segregates the 10 digits in both cases (joint and sequential). Though the absolute locations of the digit clusters differ in the two plots, the relative locations of digits share some similarity between both plots i.e. the neighboring digit clusters for each cluster are roughly similar.
This may not be sufficient to conclude that the LTM discovers the same latent representation for the underlying shared structure of tasks in these cases and we leave a more thorough investigation to future work.

\subsection{Visualizations for the jointly and sequentially learnt LTM}

\begin{figure}[ht]
\vskip 0.2in
\begin{center}
    \begin{subfigure}[b]{0.49\textwidth}
        \centering
        \includegraphics[width=0.7\textwidth]{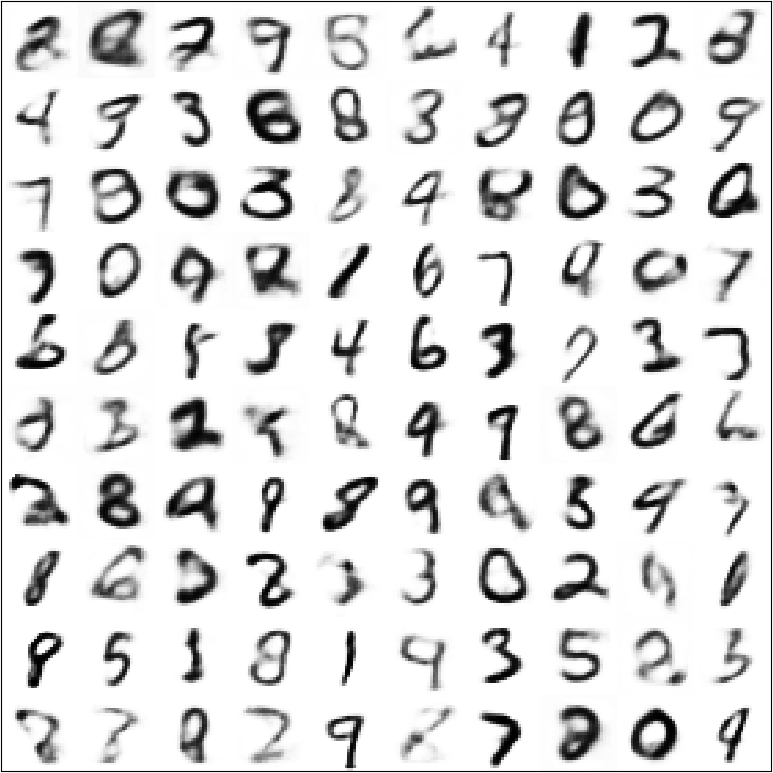}
        \caption{}
        \label{fig:joint_viz_LTM}
    \end{subfigure}
    \begin{subfigure}[b]{0.49\textwidth}
        \centering
        \includegraphics[width=0.7\textwidth]{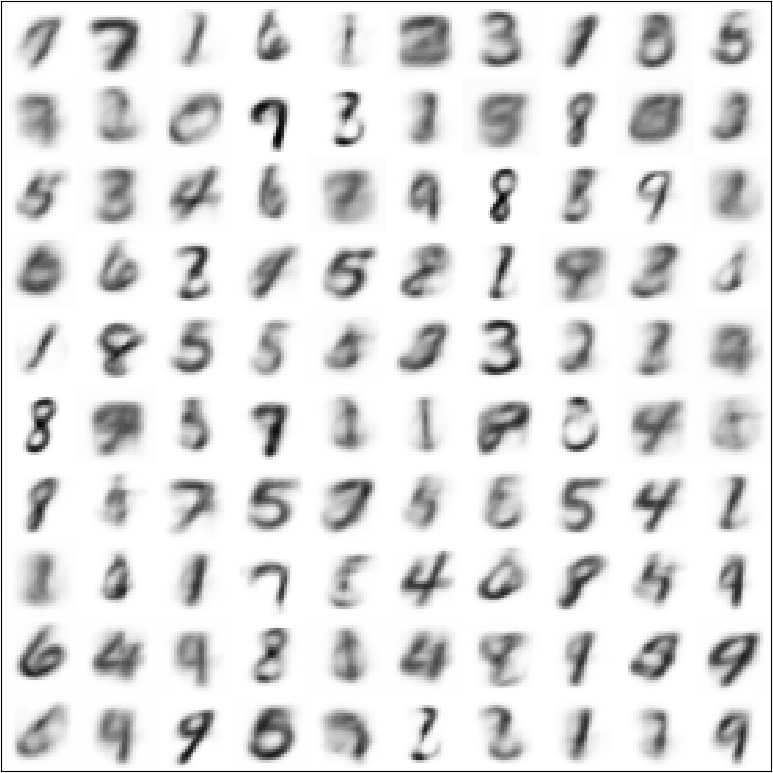}
        \caption{}
        \label{fig:seq_viz_LTM}
    \end{subfigure}
\caption{Visualization of digits from LTM when trained: (a) jointly, (b) sequentially}
\label{fig:viz_LTM}
\end{center}
\vskip -0.2in
\end{figure}

We also show visualizations of digits from the LTM when trained jointly on Digits tasks (Figure \ref{fig:joint_viz_LTM}) and when trained sequentially (Figure \ref{fig:seq_viz_LTM}). Though the digits generated from the jointly trained LTM are quite sharp, the same is not true for the sequentially trained LTM. We observe that the sequentially trained LTM produces sharp samples of the recently learnt tasks (digits $6, 7, 8$ and $9$), but blurred samples of previously learnt tasks, which is due to partial forgetting on these previous tasks.

\subsection{DGDMN with no task descriptors}

\begin{figure}[tb]
\vskip 0.2in
\begin{center}
    \begin{subfigure}[th]{0.45\textwidth}
        \includegraphics[width=\textwidth]{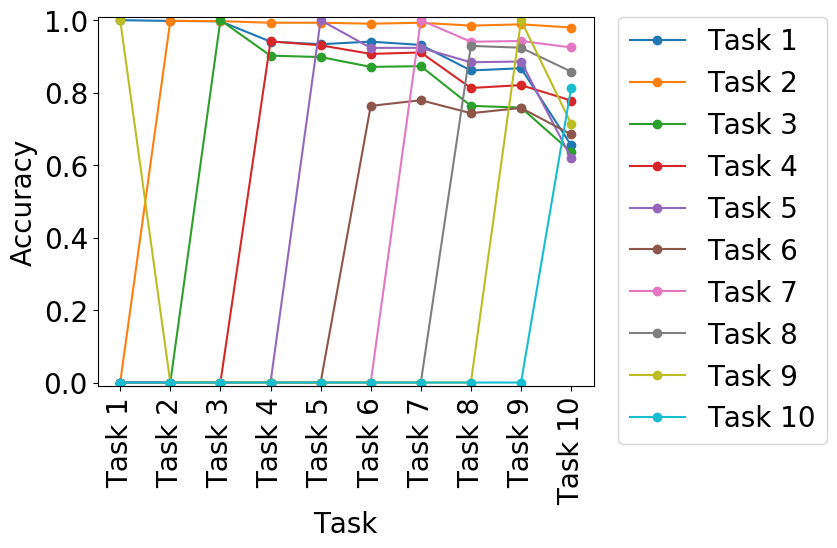}
        \caption{}
        \label{fig:digitsacc_DGDMN_recog}
    \end{subfigure}
    \begin{subfigure}[th]{0.45\textwidth}
        \includegraphics[width=\textwidth]{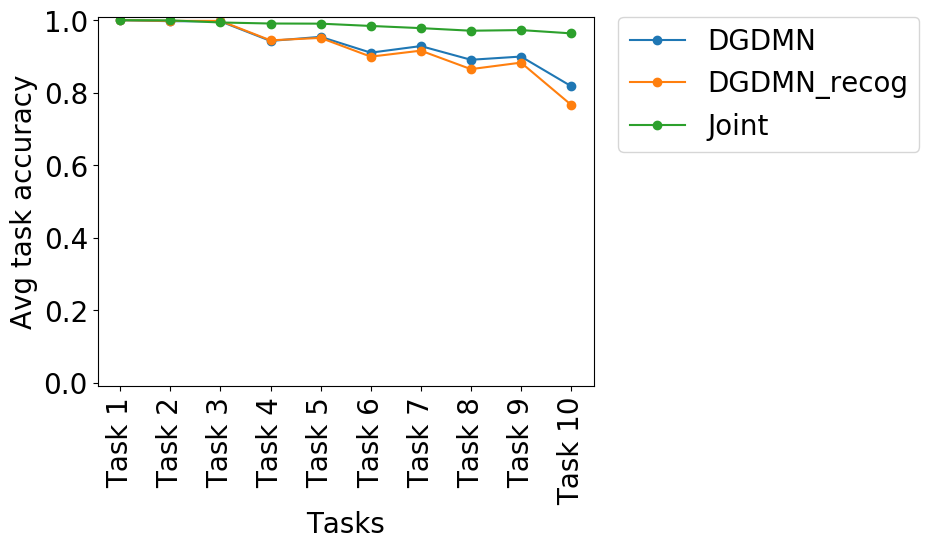}
        \caption{}
        \label{fig:digits_avgforget_DGDMN_recog}
    \end{subfigure}
\caption{Curves for DGDMN\_recog on Digits dataset: (a) Accuracy curves, (b) Average forgetting curves (x: tasks seen, y: classification accuracy).}
\label{fig:digits_DGDMN_recog}
\end{center}
\vskip -0.2in
\end{figure}

As described in section \ref{subsec:dmn}, DGDMN only uses task descriptors to recognize if a task already exists in an STTM or the LTM so that it can be appropriately allocated to the correct memory. Note that in our architecture this can also be done by using the reconstruction error of the generator on the task samples as a proxy for recognition. Specifically, in this variant DGDMN\_recog, tasks arrive sequentially but only $(X_t, Y_t)$ is observed while training and only $X_t$ while testing. A DGM, when tested to recognize task $t$ from samples $X_t$, reconstructs all samples $X_t$ using the generator $G$ and checks if the recognition loss is less than a certain threshold:
\begin{align*}
recog\_loss(X_t) = \sum_{i=1}^{N_t} \frac{\text{recons\_loss}(x_t^i)}{\text{intensity}(x_t^i)} < \gamma_{dgm},
\end{align*}
where recons\_loss($\cdot$) is the reconstruction loss on a sample, intensity($\cdot$) describes the strength of the input sample (for images, the sum of pixel intensities) and $\gamma_{dgm}$ is a scalar threshold and a hyperparameter which can be tuned separately for the LTM and the STM (same for all STTMs). We kept $\gamma_{dgm} = 1.55$ for both the LTM and all STTMs. In this case the training of the generators also employs a new termination criteria i.e. the generator of a DGM is trained till recog\_loss($\cdot$) is below $\gamma_{dgm}$.
The rest of the algorithm remains unchanged. We show the accuracy curves and the average forgetting curves for this variant on the Digits dataset in figures \ref{fig:digitsacc_DGDMN_recog} and \ref{fig:digits_avgforget_DGDMN_recog} respectively. We observe very little degradation from the original DGDMN which uses task descriptors for recognition. DGDMN\_recog achieved $ACC=0.766$ and $BWT=-0.197$ across all tasks which is similar to that of DGDMN.

%% file: src/appendixB.tex
\section{Appendix B}
\label{sec:appendixB}

\subsection{Dataset preprocessing}

All our datasets have images with intensities normalized in the range $[0.0, 1.0]$ and size $(28 \times 28)$, except Hindi which has $(32 \times 32)$ size images.

\textbf{Permnist}: Our version involved six tasks, each containing a fixed permutation on images sampled from the original MNIST dataset. We sampled $30,000$ images from the training set and all the $10,000$ test set images for each task. The tasks were as follows: (i) Original MNIST, (ii) 8x8 central patch of each image blackened, (iii) 8x8 central patch of each image whitened, (iv) 8x8 central patch of each image permuted with a fixed random permutation, (v) 12x12 central patch of each image permuted with a fixed random permutation, and (vi) mirror images of MNIST. This way each task is as hard as MNIST and the tasks share some common underlying structure.\\
\textbf{Digits}:  We introduce this smaller dataset which contains $10$ tasks with the $t^{th}$ task being classification of digit $t$ from the MNIST dataset.\\
\textbf{TDigits}: We introduced a transformed variant of MNIST containing all ten digits, their mirror images, their upside down images, and their images when reflected about the main diagonal making a total of $40$ tasks. This dataset poses similar difficulty as the Digits dataset and we use it for experiments involving longer sequence of tasks.\\
\textbf{Shapes}: This dataset was extracted from the Quick, Draw! dataset recently released by \citet{quickdraw}, which contains 50 million drawings across 345 categories of hand-drawn images. We subsampled $4,500$ training images and $500$ test images from all geometric shapes in Quick, Draw! (namely circle, hexagon, octagon, square, triangle and zigzag).\\
\textbf{Hindi}: Extracted from the Devanagri dataset~\citep{devanagari} and contains a sequence of $8$ tasks, each involving image classification of a hindi language consonant.

\subsection{Training algorithm and its parameters}

All models were trained with RMSProp~\citep{hinton2012rmsprop} using learning rate $=0.001$, $\rho=0.9$, $\epsilon=10^{-8}$ and no decay. We used a batch size of $128$ and all classifiers were provided $20$ epochs of training when trained jointly, and $6$ epochs when trained sequentially over tasks. 
For generative models (VAEs), we used gradient clipping in RMSProp with \verb|clipnorm|$=1.0$ and \verb|clipvalue|$=0.5$, and they were trained for $25$ epochs regardless of the task or dataset.

\subsection{Neural network architectures}

We chose all models by first training them jointly on all tasks in a dataset to ensure that our models had enough capacity to perform reasonably well. But we gave preference to simpler models over very high capacity models.

\textbf{Classifier Models}: Our implementation of NN, DropNN, PPR, EWC, learner for DGR and the learner for LTM in DGDMN used a neural network with three fully-connected layers with the number of units tuned differently according to the dataset ($24, 24$ units for Digits, $48, 48$ for Permnist and $36, 36$ for TDigits). DropNN also added two dropout layers, one after each hidden layer with droput rate = $0.2$ each. The classifiers (learners) for Shapes and Hindi datasets had two convolutional layers ($12, 20: 3 \times 3$ kernels for Shapes and $24, 32: 3 \times 3$ kernels for Hindi) each followed by a $2 \times 2$ max-pooling layer. The last two layers were fully-connected ($16, 6$ for Shapes and $144, 36$ for Hindi). The hidden layers used ReLU activations, the last layer had a softmax activation, and the model was trained to minimize the cross-entropy objective function. The STTM learners employed in DGDMN were smaller for speed and efficiency.

\textbf{Generative models}: The generators for DGR and LTM of DGDMN employed encoders and decoders with two fully connected hidden layers each with ReLU activation for Permnist, Digits and TDigits, and convolutional variants for Shapes and Hindi. The sizes and number of units/kernels in the layers were tuned independently for each dataset with an approximate coarse grid-search. The size of the latent variable $z$ was set to $32$ for Digits, $64$ for Permnist, $96$ for TDigits, $32$ for Shapes and $48$ for Hindi. The STTM generators in DGDMN were kept smaller for speed and efficiency concerns.

\subsection{Hyperparameters of DGDMN}

DGDMN has two new hyperparameters: (i) $\kappa$: minimum fraction of $N_{max}$ reserved for incoming tasks, and (ii) $n_{STM}$: number of STTMs (also sleep/consolidation frequency).
Both these have straightforward interpretations and can be set directly without complex hyperparameter searches.

$\kappa$ ensures continual incorporation of new tasks by guaranteeing them a minimum fraction of LTM samples during consolidation.
Given that LTM should perform well on last $K$ tasks seen in long task sequence of $T$ tasks, we observed that it is safe to assume that about $50\%$ of the LTM would be crowded by the earlier $T-K$ tasks. The remaining $0.5$ fraction should be distributed to the last $K$ tasks. So choosing $\kappa = \frac{0.5}{K}$ works well in practice (or as a good starting point for tuning). We made this choice in section \ref{subsec:longseq} with $K=10$ and $\kappa=0.05$, and hence plotted the average accuracy over the last $10$ tasks as a metric.

$n_{STM}$ controls the consolidation cycle frequency. Increasing $n_{STM}$ gives more STTMs, less frequent consolidations and hence a learning speed advantage. But this also means that fewer samples of previous tasks would participate in consolidation (due to maximum capacity $N_{max}$ of LTM), and hence more forgetting might occur. This parameter does not affect learning much till the LTM remains unsaturated (i.e. $N_{max}$ capacity is unfilled by generated + new samples) and becomes active after that. For long sequences of tasks, we found it best to keep at least $75\%$ of the total samples from previously learnt tasks to have appropriate retention. Hence, $n_{STM}$ can be set as approximately $\frac{0.25}{\kappa}$ in practice (as we did in section \ref{subsec:longseq}), or as a starting point for tuning.

\subsection{Algorithm specific hyperparameters}

\textbf{PPR}: We used a maximum memory capacity of about $3-6$ times the number of samples in a task for the dataset being learnt on (i.e. $18,000$ for Digits, $60,000$ for Permnist, $15,000$ for Shapes and $5,400$ for Hindi). While replaying, apart from the task samples, the remaining memory was filled with random samples and corresponding labels.

\textbf{EWC}: Most values of the coefficient of the Fisher Information Matrix based regularizer between $1$ to $500$ worked reasonably well for our datasets. We chose $100$ for our experiments.

\textbf{DGR and DGDMN}: $N_{max}$ for the DGM in DGR and for the LTM in DGDMN for Digits, Permnist, Shapes and Hindi was set as the total number of samples in the datasets (summed over all tasks) to ensure that there was enough capacity to regenerate the datasets well. For TDigits, we deliberately restricted memory capacity to see the effects of learning tasks over a long time and we kept $N_{max}$ as half the total number of samples. $n_{STM}$ was kept at $2$ for Digits, Permnist and Shapes, $5$ for TDigits and $2$ for Hindi. $\kappa$ was set to be small, so that it does not come into play for Digits, Permnist, Shapes and Hindi since we already provided memories with full capacity for all samples. For TDigits, we used $\kappa = 0.05$ which would let us incorporate roughly $10$ out of the $40$ tasks well.